\documentclass[sigconf]{acmart}
\usepackage[capitalize]{cleveref} 
\usepackage[htt]{hyphenat} 
\usepackage[hyphenbreaks]{breakurl}

\usepackage{subcaption}
\crefname{table}{Table}{Tables}
\usepackage{tikz}
\usetikzlibrary{positioning}
\newdimen\nodeDist
\tikzset{
	gradient/.style={fill=high!#1!low, minimum size=7mm},
	value/.style={label=center:#1, minimum size=6mm},
	botometervalue/.style={label=center:#1, minimum size=7mm},
}
\usetikzlibrary{matrix}
\newcommand{\mr}[1]{}
\newcommand{\ew}[1]{}

\newcommand{\data}[1]{\texttt{#1}}
\newcommand{\datadesc}[1]{\texttt{#1}}

\renewcommand{\paragraph}[1]{\textit{#1}}

\AtBeginDocument{%
  \providecommand\BibTeX{{%
    \normalfont B\kern-0.5em{\scshape i\kern-0.25em b}\kern-0.8em\TeX}}}

\copyrightyear{2023}
\acmYear{2023}
\setcopyright{rightsretained}
\acmConference[WWW '23]{Proceedings of the ACM Web Conference 2023}{May 1--5, 2023}{Austin, TX, USA}
\acmBooktitle{Proceedings of the ACM Web Conference 2023 (WWW '23), May 1--5, 2023, Austin, TX, USA}
\acmDOI{10.1145/3543507.3583214}
\acmISBN{978-1-4503-9416-1/23/04}

\begin{document}

\title[Simplistic Collection and Labeling Practices in Bot Detection]{Simplistic Collection and Labeling Practices Limit the Utility of Benchmark Datasets for Twitter Bot Detection}

\begin{abstract}

Accurate bot detection is necessary for the safety and integrity of online platforms. It is also crucial for research on the influence of bots in elections, the spread of misinformation, and financial market manipulation. Platforms deploy infrastructure to flag or remove automated accounts, but their tools and data are not publicly available. Thus, the public must rely on third-party bot detection. These tools employ machine learning and often achieve near-perfect performance for classification on existing datasets, suggesting bot detection is accurate, reliable and fit for use in downstream applications. We provide evidence that this is not the case and show that high performance is attributable to limitations in dataset collection and labeling rather than sophistication of the tools. Specifically, we show that simple decision rules --- shallow decision trees trained on a small number of features --- achieve near-state-of-the-art performance on most available datasets and that bot detection datasets, even when combined together, do not generalize well to out-of-sample datasets. Our findings reveal that predictions are highly dependent on each dataset's collection and labeling procedures rather than fundamental differences between bots and humans. These results have important implications for both transparency in sampling and labeling procedures and potential biases in research using existing bot detection tools for pre-processing.

\end{abstract}



\author{Chris Hays}
\authornote{Both authors contributed equally to this research.}
\affiliation{%
    \institution{Massachusetts Institute of Technology}
    \country{}
    \thanks{Code for the results presented is available on GitHub at \href{https://github.com/johnchrishays/bot-detection}{github.com/johnchrishays/bot-detection} and can be accessed with DOI \href{https://doi.org/10.5281/zenodo.7196519}{doi.org/10.5281/zenodo.7196519}. The authors would like to thank {Sinan Aral, Dean Eckles, and Kiran Garimella for helpful research discussions, as well as the anonymous reviewers for helping improve the presentation of the results. E.W.\ is supported by the NSF Graduate Research Fellowship Program.}}
}
\email{jhays@mit.edu}

\author{Zachary Schutzman}
\authornotemark[1]
\affiliation{%
    \institution{Massachusetts Institute of Technology}
    \country{}
}
\email{zis@mit.edu}

\author{Manish Raghavan}
\affiliation{%
    \institution{Massachusetts Institute of Technology}
    \country{}
}
\email{mragh@mit.edu}

\author{Erin Walk}
\affiliation{%
    \institution{Massachusetts Institute of Technology}
    \country{}
}
\email{ewalk@mit.edu}

\author{Philipp Zimmer}
\affiliation{%
    \institution{Massachusetts Institute of Technology}
    \country{}
}
\email{philippz@mit.edu}

\maketitle

\section{Introduction}
\label{sec:intro}

%

With the rise of online social media as an important means for connecting with others and sharing information, the influence of \textit{bots}, or automated accounts, has become a topic of vital societal concern.
%
%
Some bots are benign and serve content which is entertaining or directly enhances the accessibility of the site (e.g., by providing captions for otherwise uncaptioned videos on a platform), but many others engage in influence operations, the spread of misinformation, and harassment: 
fake followers boost some users' perceived popularity; spammers flood the site with advertisements for a political candidate or product; malicious automated accounts undermine the credibility of elections or inflame polarization.
Bots have reportedly influenced the 2016 US Presidential Election \cite{bessi2016social, gorodnichenko2018social}, the Brexit vote in the UK \cite{bastos2019brexit, gorodnichenko2018social}, the spread of misinformation about COVID-19 \cite{ferrara2020covid} and financial markets \cite{nizzoli2020access, cresci2019fake}.  
The ability (or inability) to accurately label such accounts could have a very real impact on elections and public health as well as public trust in institutions.

%
Platforms remove large numbers of accounts that they deem inauthentic, but they keep these removal systems secret and may be incentivized to misrepresent the influence or prevalence of bots.
Indeed, bot detection was at the center of Elon Musk's negotiations to buy Twitter:
Twitter claimed that less than five percent of its monetizable users are bots \cite{twitter2021mdau} while Musk claimed the number is much higher \cite{musk2022mdau}. 
%
%
Because internal bot detection techniques are in general not made public, researchers, journalists, and the public at-large rely on researcher-developed tools to separate bots from genuine human users and understand the impact of bots on social phenomena.

Developing tools for bot detection on Twitter {and other online social media platforms} is an active area of research.
Over the last decade, an abundance of user datasets have been collected for the purpose of enabling third-party bot detection.
Tools trained on these datasets achieve high (sometimes nearly perfect) performance using expressive machine learning techniques such as ensembles of random forests and deep neural networks, and hundreds or thousands of features such as profile metadata, engagement patterns, network characteristics, and tweet content and sentiment.
%

%
Crucially, researchers frequently use bot detection as a \textit{preprocessing} step to study social phenomena, to separate human users from bots and study phenomena related to one or both of humans and bots.
This includes topic areas such as the spread of mis- or disinformation \cite{jang2018a, vosoughi2018spread, shao2018the, shu2020fakenewsnet:, pennycook2021shifting, shao2018anatomy, broniatowski2018}, elections \cite{badawy2018analyzing, bessi2016social, ferrara2017disinformation, keller2018social, pierri2020investigating, stella2018bots} and echo chambers \cite{choi2020rumor} and published in premier venues for scientific research including \textit{Science} \cite{vosoughi2018spread}, \textit{Nature} \cite{pennycook2021shifting} and \textit{PNAS} \cite{stella2018bots}. 
For example, \citet{broniatowski2018} observed that bots eroded the population's trust in vaccinations, \citet{gonzalezbailon2022} conclude bots share disproportionate amount of content during political protests and \citet{vosoughi2018spread} conclude that humans and bots spread fake news in different ways. 
%
The robustness and validity of these results depend on accurate and reliable bot detection.
%
%

Third-party bot detection tools are also easily accessible to and widely used by the public: the most recent version of Botometer \cite{sayyadiharikandeh2020detection} reportedly receives hundreds of thousands of daily queries to its public API \cite{yang2022botometer101} and BotSentinel \cite{botsent_2022} provides a browser extension and ways to conveniently block accounts classified as bots. 

\paragraph{Is bot detection {fit for downstream use}?}
{To an outsider --- someone who might want to use bot detection but has not done research on the topic --- bot detection might seem like a case study in the successful application of machine learning to an important problem}: researchers have collected a variety of datasets for a well-defined classification task and expressive machine learning models like random forests and neural networks attain near-perfect performance on the data. 
Moreover, these methods have been widely adopted in both the academic literature and in public use.
Bot detection tools are frequently trained on a combination of datasets, and researchers have argued that the existing approach can adapt to the short-comings of existing classifiers or evolution of more human-like bots by adding more datasets \cite{sayyadiharikandeh2020detection} or using even more complex techniques, like generative adversarial networks \cite{cresci2020decade}.

Even so, there are signs that bot detection tools are far from perfect. They may disagree with one another~\cite{martini2021bot}, prove unreliable over time~\cite{rauchfleisch2020false}, and rely on dubious labels~\cite{gallwitz2021therise,gallwitz2022investigating}. 
{Moreover, bot detection researchers certainly do not consider the problem solved and have, in previous research, observed that bot classifiers can fail to generalize \cite{echeverria2018lobo} and surfaced concerns that more sophisticated bots may go undetected \cite{cresci2020decade}.}
Here, we attempt to reconcile and systematically explain the apparent achievements of Twitter bot detection with what seem to be significant challenges and limitations.

Evaluating third-party bot detection datasets and tools is inherently challenging: the ``ground truth'' is unknown or inaccessible to the public, and the only window of insight we have into bots on Twitter is through the datasets themselves.
However, this does not make evaluation impossible.
We can still gain a better understanding of what these datasets tell us by closely analyzing the datasets and how they relate to one another.

\begin{figure}
\centering
    \begin{minipage}{.29\columnwidth}
\begin{tikzpicture}[
    node/.style={%
      draw,
      rectangle,
      font=\fontsize{7pt}{7pt}\selectfont
    },
  ]

    \node [node] (A) [draw, align=center, font=\fontsize{7pt}{7pt}\selectfont] {Ever tweeted  the \\ word `earthquake'?};
    \path (A) ++(-105:\nodeDist) node [node] (B) {bot};
    \path (A) ++(-75:\nodeDist) node [node] (C) {human};

    \draw (A) -- (B) node [left,pos=0.5,font=\fontsize{7pt}{7pt}\selectfont] {NO}(A);
    \draw (A) -- (C) node [right,pos=0.5,font=\fontsize{7pt}{7pt}\selectfont] {YES}(A);
\end{tikzpicture}
\end{minipage}
\hfill
\begin{minipage}{.29\columnwidth}
\begin{tikzpicture}[
    node/.style={%
      draw,
      rectangle,
      font=\fontsize{7pt}{7pt}\selectfont
    },
  ]

    \node [node] (A) [draw, align=center] {Liked more than\\ 16 tweets?};
    \path (A) ++(-105:\nodeDist) node [node] (B) {bot};
    \path (A) ++(-75:\nodeDist) node [node] (C) {human};

    \draw (A) -- (B) node [left,pos=0.5,font=\fontsize{7pt}{7pt}\selectfont] {NO}(A);
    \draw (A) -- (C) node [right,pos=0.5,font=\fontsize{7pt}{7pt}\selectfont] {YES}(A);

\end{tikzpicture}
\end{minipage}
\hfill
\begin{minipage}{.39\columnwidth}
\begin{tikzpicture}[
    node/.style={%
      draw,
      rectangle,
      font=\fontsize{7pt}{7pt}\selectfont
    },
  ]

    \node [node] (A) [draw, align=center]{Account created\\ after Nov.\ 27, 2009?};
    \path (A) ++(-110:\nodeDist) node [node] (B) [draw, align=center] {bot};
    \path (A) ++(-70:\nodeDist) node [node] (C) [draw, align=center] {Follow more than\\ 508 accounts?};
    \path (C) ++(-105:\nodeDist) node [node] (D) {human};
    \path (C) ++(-75:\nodeDist) node [node] (E) {bot};

    \draw (A) -- (B) node [left,pos=0.5,font=\fontsize{7pt}{7pt}\selectfont] {NO}(A);
    \draw (A) -- (C) node [right,pos=0.5,font=\fontsize{7pt}{7pt}\selectfont] {YES}(A);
    \draw (C) -- (D) node [left,pos=0.5,font=\fontsize{7pt}{7pt}\selectfont] {NO}(A);
    \draw (C) -- (E) node [right,pos=0.5,font=\fontsize{7pt}{7pt}\selectfont] {YES}(A);
\end{tikzpicture}
\end{minipage}
    \caption{Two shallow decision trees for \data{cresci-2017} (left, middle) achieving accuracies of 0.98 and one for \data{caverlee-2011} (right) with an accuracy of 0.91.}   %
    \label{fig:example-trees}

\end{figure}
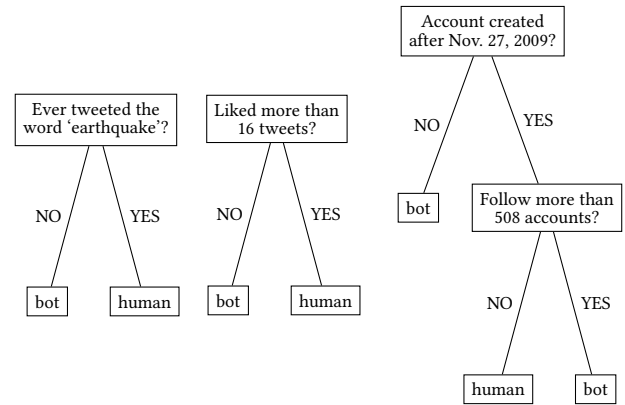

Consider, for example, the dataset released by \citet{cresci2017the} (\data{cresci-2017}), one of the most widely used in the academic literature. This dataset consists of a pool of genuine human users, collections of fake followers, and several types of `spam bots': a diverse collection of accounts in this domain.  
The state-of-the-art model is a deep neural network using text data which achieves essentially perfect performance on this dataset \cite{kudugunta2018deep}.  
However, a closer inspection revealed something surprising: we can achieve near-state-of-the-art performance using a classifier that asks a single yes/no question of the data. 
In fact, there are at least two different yes/no questions which nearly separate the human and bot classes. 
These classifiers are shown in the left and middle decision trees in \Cref{fig:example-trees}.
The left tree is an artifact of convenience sampling from {\citet{avvenuti2017hybrid}}, which concerns social sensing of natural disasters using Twitter.\footnote{{A previous version of this paper speculated that the data might have been from a project related to \citet{cresci2015linguistically}'s work on the social sensing of earthquakes.  \Citet{cresci2023demystifying} have since clarified that the data originated in the work of \citet{avvenuti2017hybrid}, which is on a similar topic and shares authors in common.}}
%
%
On the right of \Cref{fig:example-trees} we show another high-performing classifier for another popular dataset: \data{caverlee-2011} published in \cite{lee2011a}. 
%
%
Again, a small number of yes/no questions distinguishes humans from bots with high accuracy.
These examples are not exceptional cases. 
%
%
As we will show, almost all of the other benchmark datasets we analyze admit high performance using very simple classifiers. 

How should we reconcile these results with our intuition that that bot detection is a difficult problem?
On the one hand, it is possible that bot detection is inherently simpler than expected, and simple decision rules suffice.
On the other, perhaps the datasets themselves fail to capture anywhere near the true complexity of bot detection.
If this is the case, then while simple decision rules perform well in-sample, their performance will be significantly worse when deployed.
We provide evidence to support the latter hypothesis across a wide range of Twitter bot detection datasets.

\paragraph{Our contributions.} 
In this work, we carefully examine widely used datasets for Twitter bot detection and explore their limitations. 
First, we demonstrate that simple decision rules perform nearly as well as the state-of-the-art models on benchmark datasets. 
Thus, each dataset only provides predictive signals of limited complexity.
Because our simple decision rules allow us to transparently inspect the reasons for our classifiers' high performance, we find that predictive signals in the datasets likely reflect particular collection and labeling procedures, i.e., the processes for collecting accounts from Twitter and assigning a human or bot label to each account. 

Next, we examine combinations of datasets.
Many bot detection tools combine datasets (see, e.g., \cite{dimitriadis2021social, guo2022social, yang2020scalable}) and argue implicitly or explicitly that it is possible to cover the distribution of bots that appear on Twitter by doing so. 
Building on prior work \cite{sayyadiharikandeh2020detection,echeverria2018lobo}, we show that expressive machine learning models trained on one dataset do not perform well when tested on others and that models trained on all but one dataset perform poorly when evaluated on the held-out one.
Information provided by a dataset does not generalize to others, suggesting that datasets are distributed according to dissimilar distributions, which indicates different sampling (i.e.,  collection and labeling) procedures.
%

Finally, we consider whether imposing structural assumptions about the data, namely that each dataset contains bots from one of a small number of \textit{types} (e.g., spam bots or fake followers) can yield greater generalization as the approaches in \citet{sayyadiharikandeh2020detection} and \citet{dimitriadis2021social} suggest. 
We find that simple decision rules can accurately differentiate bots of each type from humans.
Thus, each sample of bots of one type is itself of low informational complexity.
We additionally show that, within accounts of a particular bot type, simple decision rules can identify from which dataset a given bot originates.
Thus, datasets of a given bot type are drawn from very different distributions, again indicating different data collection procedures. Taken together, these results suggest that each individual dataset contains little information, predictive signals in each dataset are not informative for prediction on the others, and this is true even within datasets representing a particular type of bots.
Therefore, existing datasets are unlikely to provide a representative or comprehensive sample of bots and {our analysis can explain \textit{why}} it is unlikely that classifiers trained on this data will perform well when deployed.

Beyond bot detection, our methodology --- examining simple decision rules on datasets and measuring cross-dataset performance --- may be useful for detecting simplistic data sampling and labeling processes in a range of machine learning applications: If datasets admit highly accurate simple decision rules, the datasets themselves have low informational complexity. 
If, additionally, expressive machine learning models trained on some datasets do not generalize to other datasets, the underlying system does not appear to be simple, and the datasets are unlikely to provide insight into the problem domain as a whole.

We also believe these findings have direct implications for future bot detection research {both on Twitter and beyond}: creators of bot detection datasets should transparently report and justify sampling and labeling procedures; researchers developing bot detection techniques should train and analyze simple, interpretable models alongside more expressive ones; and researchers using bot detection as a preprocessing step should consider how it may bias results.
\section{Background}
\label{sec:background}
\paragraph{Bot detection techniques.} 
%
%
%
%
%
Researchers have used a range of cutting-edge machine learning techniques to bot detection across diverse types of data in order to improve classification. 
One approach applies random forests \cite{yang2013empirical, gilani2017classification} and ensembles of random forests that combine predictions from classifiers trained on subsets of data \cite{sayyadiharikandeh2020detection,dimitriadis2021social}.
Another popular approach leverages text data to apply large pre-trained language models \cite{heidari2020using} or models trained by the researchers themselves \cite{ilias2021detecting, luo2019deepbot, garcia-silva2019an, kudugunta2018deep, ilias2021deep}.  
A third approach uses network data to train graph neural networks \cite{feng2022heterogeneity-aware, alothali2022bot-mgat:,feng2021botrgcn} or try to detect botnets from anomolous network structures~\cite{wu2017adaptive}.
Finally, a fourth approach seeks insight from other disciplines by using behavioral \cite{gilani2019a, giorgi2021characterizing} or biology-inspired techniques \cite{cresci2018social, cresci2016dna-inspired, rovito2022an, cresci2017exploiting}.
In addition to novel predictive models, significant effort is spent deriving or exploring profile, text or network features that are likely to be informative for bot detection \cite{ilias2021detecting, mazza2019rtbust}.
All of the papers cited above rely on the benchmark datasets analyzed in our work.
%

        

\paragraph{Limitations of bot detection tools.} 
Several papers have explored the limitations of bot detection techniques, but few provide evidence to explain these limitations. To the best of our knoweldge, our work is the first that traces the limitations of bot detection to simplistic sampling and labeling strategies. 
\citet{martini2021bot} compare three public tools for bot detection and finds significant disagreement of predictions across tools. 
Relatedly, \citet{rauchfleisch2020false} find that a single tool may produce varying results over time as a result of variation in account activity and \citet{torusdag2020secure} created bots that can reliably evade existing bot detection frameworks.
\citet{elmas2022characterizing} find that the qualitative observations of prior work, such as that bot accounts are typically recently created or are marked by a high volume of activity, do not hold on data collected for their paper and conclude that popular classifiers may not generalize.
%
%
\citet{gallwitz2021therise, gallwitz2022investigating} manually identify individual accounts that are incorrectly labeled as `bots' in popular datasets, noting a high prevalence of false positives and arguing that the labels which are typically taken as ground truth may have errors. 
{Researchers have also explored and tried to quantify the practical difficulties of bot detection. For example, \citet{cresci2020decade} argues that bots may become more sophisticated over time to evade detection and \citet{echeverria2018lobo} provide evidence to suggest that existing tools may not generalize to out-of-sample data. By contrast, our analysis in this work focuses on \textit{finding concrete, quantifiable explanations} for these limitations, developing a framework for evaluating whether datasets are suitable to the prediction task at hand, and providing recommendations for future best practices in data collection and labeling.}

\section{Data \& Methods}
\label{sec:datasets}

In this section, we discuss the datasets we analyze and our criteria for including each in our analysis. 
Most benchmark datasets in the literature are aggregations of data collected across various contexts. The benchmark datasets we study are summarized in \Cref{tab:agg-datasets}.

\subsection{Dataset collection.} To collect a list of benchmark datasets, we searched Google Scholar for peer-reviewed papers related to bot detection and within the references of papers we found. 
%
%
We found a total of 58 papers using at least one of the datasets we included in our analysis, of which 22 had at least 50 citations on Google Scholar at the time of writing (while several had at least 500 citations) and 26 of which were published since 2020.
In our analysis, we only include datasets that were used in multiple peer-reviewed bot detection papers reporting accuracy and F1 scores that we found in our search, although nearly all of the datasets were used in many more than two.
Several of the datasets were accessed via the the Botometer Bot Repository.\footnote{https://botometer.osome.iu.edu/bot-repository/datasets.html}
For the rest of the datasets, we reached out to the author(s) of the associated paper to request access to the original data (for \data{twibot\hyp{}2020} and \data{yang\hyp{}2013}) or found public access to the data online (in the cases of \data{caverlee\hyp{}2011} and \data{pan\hyp{}2019}). %

%
We also received augmented data for \data{gilani\hyp{}2017} as was used in the original work \cite{gilani2019a, gilani2017of, gilani2017classification} from the authors, though a reduced feature set is available on the Bot Repository. 
For \data{gilani\hyp{}2017} and \data{caverlee\hyp{}2011}, the original data provided by the authors \cite{lee2011a, gilani2017classification} contained at least 35\% more users than are included in the Bot Repository; we use the larger, original datasets in our results.
For the \data{astroturf} and \data{varol\hyp{}2017} datasets published on the Bot Repository, the data only came as a list of user identifiers. Due to the amount of time that has passed since their origination, we did not rehydrate that data or use it in our analysis.

\paragraph{Features.} All of the datasets include profile characteristics, which typically include screen name, number of tweets, number of followers, number of following, number of favorites, language, location, timezone, number of Twitter lists on which the user is included, and others. 
Additionally, several datasets include a corpus of tweets by each of the users in the dataset. 
Network relations and associated following/followers behavior are occasionally recorded.
%

%
\paragraph{Annotation methods.} 
Ascertaining `ground truth' labels for bot detection is a challenging task. In most datasets, humans, either the authors of the paper or hired crowdworkers, assigned a `bot' or `human' label to each account manually. 
Previous work has found human annotators have a high level of agreement with each other \cite{gilani2017classification}, and accounts for which there is not enough agreement are sometimes excluded from the datasets \cite{feng2021twibot}. 
Others used heuristics or relied on external sources (e.g., celebrity accounts [\data{celebrity\hyp{}2019}] or accounts that tweeted links from public blacklists [\data{yang\hyp{}2013}]) to assign them. 
The quality of the hand- and heuristic-labeled datasets depends crucially on the implicit assumption that \textit{humans} are very good at the classification task, and neither the datasets themselves nor the broader literature provide robust evidence that this is the case. 
To the contrary, recent evidence suggests human annotators are systematically biased towards believing opinion-incongruent accounts are bots \cite{yan2021asymmetrical, wischnewski2021disagree}. %
Similarly, there are accounts for which neither bot nor human labels may be appropriate, such as semi-automated accounts or accounts that represent an institutional entity like a corporation or a university \cite{chu2012detecting}.
Nevertheless, since other work assumes labels in the data are ground truth and since better annotation methods are not available, we make the same assumption.

\subsection{Dataset descriptions.}
The datasets which we considered fall into two categories: component datasets, which consist of a single class (human or bot) of accounts, and composite datasets, which consist of a combination of component datasets. Each of the 28 datasets is described briefly below. Unless otherwise specified, the authors of the associated paper hand-labeled the dataset.

\datadesc{\data{social\hyp{}spambots\hyp{}1}} \cite{cresci2017the} are spam accounts used during the 2014 Roman mayoral election to promote a particular candidate.
\datadesc{\data{social\hyp{}spambots\hyp{}2}} \cite{cresci2017the}  are spammers who promoted the Talnts app using the hashtag \#TALNTS.
\datadesc{\data{social\hyp{}spambots\hyp{}3}} \cite{cresci2017the}  contains accounts which spammed links to products on Amazon, both genuine links to products as well as malicious URLs. 
\datadesc{\data{traditional\hyp{}spambots\hyp{}yang}} \cite{yang2013empirical} are accounts spamming known malicious links, collected by crawling the Twitter network. 
\datadesc{\data{genuine\hyp{}accounts\hyp{}yang}} \cite{yang2013empirical} are accounts which did not tweet a malicious link, taken from the same crawling process as \data{traditional\hyp{}spambots\hyp{}yang}.
\datadesc{\data{traditional\hyp{}spambots\hyp{}2}} \cite{cresci2017the} includes accounts that share malicious URLs and accounts that repeatedly tag those sharing such content. 
\datadesc{\data{traditional\hyp{}spambots\hyp{}3}} \cite{cresci2017the} and \datadesc{\data{traditional\hyp{}spambots\hyp{}4}} \cite{cresci2017the} are accounts spamming job offers. 
\datadesc{\data{pronbots\hyp{}2019}} \cite{yang2019arming} are Twitter bots infrequently tweeting links to pornographic sites.
\datadesc{\data{elezioni\hyp{}2015}} \cite{cresci2015fame}  are manually labeled Italian language accounts that used the hashtag \#elezioni2013.  
\datadesc{\data{political\hyp{}bots\hyp{}2019}} \cite{yang2019arming} were collected and identified by Josh Russell (@josh\_emerson) to be automated accounts run by a single individual to amplify right-wing influence in the U.S. 
%
\datadesc{\data{midterm\hyp{}2018}} \cite{yang2020scalable} includes accounts which used relevant hashtags such as \#2018midterms during the 2018 US elections. 
\datadesc{\data{stock\hyp{}2018}} \cite{cresci2019fake}  are accounts with high volumes of tweets that tagged stock microblogs with cashtags.
\datadesc{\data{genuine\hyp{}accounts\hyp{}cresci}} \cite{cresci2017the} is purportedly a random sample of human Twitter users, confirmed to be genuine by their response to a natural language question. These are the accounts that all tweeted ``earthquake''{, using data from previous work on crowdsensing for natural disasters \cite{avvenuti2017hybrid}}, mentioned in \Cref{sec:intro} and discussed in \Cref{sec:empirical}.
\datadesc{\data{twibot\hyp{}2020}} \cite{feng2021twibot} was collected by crawling the Twitter network using well-known users as seeds. The accounts were manually labeled by hired crowdworkers. 
%
\datadesc{\data{gilani\hyp{}2017}} \cite{gilani2017classification} contains accounts sampled from Twitter's streaming API. 

%
\datadesc{\data{rtbust\hyp{}2019}} \cite{mazza2019rtbust} contains manually labeled accounts subsampled from all accounts which retweeted Italian tweets during the data collection period.
\datadesc{\data{fake\hyp{}followers\hyp{}2015}} \cite{cresci2017the} and \datadesc{\data{vendor\hyp{}purchased\hyp{}2019}} \cite{yang2019arming} are {fake follower} accounts purchased from different Twitter online markets.
\datadesc{\data{caverlee\hyp{}2011}} \cite{lee2011a} were  collected via honeypot Twitter accounts and researchers used a human-in-the-loop automated process to label bot and human accounts.
\datadesc{\data{celebrity\hyp{}2019}} \cite{yang2019arming} are manually collected verified celebrity accounts.
\datadesc{\data{the\hyp{}fake\hyp{}project\hyp{}2015}} \cite{cresci2015fame} consists of accounts which followed  @TheFakeProject and successfully completed a CAPTCHA.
\datadesc{\data{botwiki\hyp{}2019}} \cite{yang2020scalable} is a list of self-identified benign Twitter bots, for example automated accounts that post generative art or tweet world holidays.
\datadesc{\data{feedback\hyp{}2019}} \cite{yang2019arming} is a collection of about 500 accounts which users of Botometer flagged as being incorrectly labeled by that tool.

Several of the datasets we study are combinations of the above components. 
\datadesc{\data{cresci\hyp{}2015}} \cite{cresci2015fame} includes \data{the\hyp{}fake\hyp{}project\hyp{}2015},\   \data{elezioni\hyp{}2015}, and \data{fake\hyp{}followers\hyp{}2015}. 
\datadesc{\data{cresci\hyp{}2017}} \cite{cresci2017the} is composed of \data{fake\hyp{}followers\hyp{}2015}, \data{genuine\hyp{}accounts\hyp{}cresci}, the three \data{social\hyp{}spambots} datasets, and the four \data{traditional\hyp{}spambots} datasets.
\datadesc{\data{yang\hyp{}2013}} \cite{yang2013empirical} has bots from \data{traditional\hyp{}spambots\hyp{}yang} and humans from \data{genuine\hyp{}accounts\hyp{}yang}.
\datadesc{\data{pan\hyp{}2019}} \cite{rangel2015overview} includes all of the components of \data{cresci\hyp{}2015}, \data{cresci\hyp{}2017}, \data{varol\hyp{}2017}, plus \data{caverlee\hyp{}2011} and an additional collection of manually annotated bots and humans not found in any of these. This dataset additionally includes tweet data not present in the original components.

    \begin{table}[]

\caption{Publicly available benchmark datasets for bot detection.}\small
    \begin{tabular}{l r r r} 
    \toprule 
    Benchmark dataset & Ref. & Humans & Bots \\
    \midrule
\data{twibot-2020} & \cite{feng2021twibot} & 3632 & 4646 \\
\data{feedback-2019} & \cite{yang2019arming} & 380 & 139 \\
\data{pan-2019} & \cite{rangel2015overview} & 2060 & 2060 \\
\data{rtbust-2019} & \cite{mazza2019rtbust} & 340 & 353 \\
\data{midterm-2018} & \cite{yang2020scalable} & 8092 & 42446 \\
\data{stock-2018} & \cite{cresci2019fake} & 6174 & 7102 \\
\data{cresci-2017} & \cite{cresci2017the} & 3474 & 10894 \\
\data{gilani-2017} & \cite{gilani2017classification} & 1939 & 1492 \\
\data{cresci-2015} & \cite{cresci2015fame} & 1957 & 3351 \\
\data{yang-2013} & \cite{yang2013empirical} & 10000 & 1000 \\
\data{caverlee-2011} & \cite{lee2011a} & 19276 & 22223 \\

    \end{tabular}
    \label{tab:agg-datasets}
\end{table}


           
        

%


%


\subsection{Methods.}

\paragraph{Simple decision rules.} 
While sophisticated machine learning models are able to learn complicated relationships between patterns in input data and their labels, their flexibility generally comes at the cost of transparency and interpretability.  
We choose to instantiate `simple decision rules' as shallow decision trees because their transparency allows us to easily examine exactly why each data point is assigned a label. 
%
%
Similar analyses are significantly more difficult or infeasible with the complex and opaque models predominantly used in bot detection. 
Researchers have used now-standard explainable machine learning tools like LIME \cite{ribeiro2016why} and SHAP \cite{lundberg2017a} to bot detection models \cite{cresci2015fame, kouvela2020bot, yang2020scalable}.
However, none of these can demonstrate as we do that the underlying datasets admit simple, high-performing classifiers that rely on a small number of features. 
Other simple machine learning models like linear regression, $k$-means, or a nearest neighbors classifier may be able to provide similar interpretability to shallow decision trees, but the choice of the exact method is not important for our analysis.

We use \texttt{scikit\hyp{}learn}'s implementation of binary decision  trees,\footnote{\url{https://scikit-learn.org/stable/modules/tree.html}} which are trained recursively on numerical data by choosing a feature-threshold pair (represented by a node) that best splits the data into two groups by class and then learning a decision tree on each group separately. 
{When we incorporated text data (as in the left-most tree in \cref{fig:example-trees}), we used one-hot encoding for each word in the corpus.}
%
%
In our case, after a fixed recursive depth (corresponding to tree depth), the classifier outputs a label corresponding to that of the majority of examples in the group; these are leaves of the tree. 
We only consider trees of depth four or less to ensure the trees can be readily inspected and to avoid overfitting.  
%
%
%
See \Cref{fig:example-trees} for several examples of shallow decision trees trained on benchmark datasets.

\paragraph{Performance metrics.} 
The most commonly reported metrics used in the literature are accuracy and the F1 score. 
%
{{Accuracy} is defined as the fraction of examples labeled correctly. 
When a dataset is not balanced between classes, the accuracy may be misleading since a naive model can achieve a high accuracy by always predicting the majority class.} 
{The {F1 score} in binary classification is the harmonic mean of the model's precision and recall. 
In our context, low F1 score indicates a classifier that either does not detect a high proportion of the bots or incorrectly labels a large fraction of the humans. 
The F1 score does not incorporate the number of true negatives, i.e., humans correctly labeled as humans, which is potentially misleading in contexts where bots outnumber humans. }

Though the two metrics complement each other, both depend on the proportions of humans and bots in the data. 
For these reasons, it is hard to compare accuracy and F1 score results across models and datasets with different proportions of bots and humans. To provide additional clarity and comparability, we report the {balanced accuracies} (bal.\ acc.) of our classifiers, or the arithmetic mean of the true positive rate and the true negative rate. Balanced accuracy is a less useful metric when one has prior knowledge about the relative proportions of bots and humans in the context where the classifier will be deployed. 

\section{Results}
\label{sec:empirical}

%

In this section we present and discuss the results of experiments run on the datasets identified in \Cref{sec:datasets}. 
In \Cref{subsec:individual-benchmarks}, we establish that simple decision rules, instantiated as shallow decision trees, yield near-state-of-the-art performance when trained and evaluated on these benchmark datasets and that the simple decision rules are suggestive of sampling and labeling procedures. 
%
In \cref{subsec:cross-dataset-generalization}, we show that the information contained in one dataset is not informative for classification on other datasets, in other words classifiers trained on one dataset do not generalize to other datasets.  
Building on prior work~\cite{echeverria2018lobo}, we next establish that training a classifier on all of the datasets but one and testing on the held out one yields performance not much better than random guessing. 
Predictably, both of these results are weaker in the cases where the held out dataset shares some data with the training dataset(s). 

In \cref{subsec:bot-types}, we assess an assumption made among popular bot detection tools in the literature \cite{echeverria2018lobo, sayyadiharikandeh2020detection, dimitriadis2021social}: each dataset of bots represents one of a small number of types of bots, like spam bots or fake followers. 
This assumption underpins the approach of training a series of specialized classifiers to detect each bot type and then combining their outputs to provide an overall prediction.
We find it is indeed possible to build simple classifiers that perform well on differentiating one type of bot from humans consistent with prior work using more sophisticated models \cite{sayyadiharikandeh2020detection}.

However, in \Cref{subsec:distinguishing-bot-types}, we provide evidence that datasets within a given bot type are not drawn from similar distributions; simple decision rules can also differentiate bots within the same type from different datasets.
This implies that rather than each dataset being broadly representative of all or a part of the respective subspace containing that type of bot, these component datasets are drawn from narrow and easily separable regions of the sample space, meaning that the signals from each dataset are strongly influenced by sampling and labeling procedures.
We conclude that we should not expect that the collection of more --- even many more --- datasets using similar simplistic sampling and labeling strategies will result in significantly more generalizable classifiers.

\subsection{Decision trees on component datasets.} \label{subsec:individual-benchmarks} 
In \Cref{table:sdt-summary}, we summarize the performance of our decision trees against that of the state-of-the-art classifier on each dataset. 
For each dataset, we train a tree of each depth from one to four and report the accuracy and F1 score for the shallowest tree that achieves test accuracy and F1 score within 2.5 percent of best-performing tree; in other words, we favor shallower trees when performance is similar across depths. 

In training and testing our models, we use five\hyp{}fold cross\hyp{}validation and report the results accordingly. However, \data{twibot\hyp{}2020} comes with a train/test split, which we use instead for comparability with prior work.
We searched the literature to find the state-of-the-art performance on each dataset. %
In the cases of \data{midterm-2018} and \data{stock-2018}, we could not find papers reporting results on these datasets alone, so we omit entries for the state-of-the-art from the table. (In the literature, e.g., \cite{guo2022social, dimitriadis2021social, wang2021detecting, sayyadiharikandeh2020detection, mou2020malicious}, these datasets are frequently used in combination with other ones.) 
Where a paper reported the results from multiple models or from different test sets, we recorded the maximum score achieved for each metric to make our analysis as conservative as possible.

{We only train our shallow decision trees on the types of features used in the state-of-the-art models in order to make our analysis as conservative as possible.} 
Thus, if a classifier from the literature is trained on profile features like the number of followers or number of tweets, but not text features, we use only profile features in our model.
In many cases, the full set of features that were used to develop state-of-the-art models were not publicly available. 
Despite this, for almost all of the datasets, our simple decision rules perform nearly as well as the state-of-the-art. 
For all datasets except \data{rtbust\hyp{}2019}, accuracy is within ten percentage points of the state of the art and all of those but \data{gilani\hyp{}2017} and \data{caverlee\hyp{}2011} are within five percentage points of the state of the art.
For most datasets, the F1 score is similarly close to the state of the art.
%

For the ``earthquake'' classifier on \data{cresci-2017} described in \Cref{sec:intro,fig:example-trees}, an inspection of several of the human-labeled accounts on Twitter after we had trained the simple model revealed that they tweeted the word ``earthquake,'' after which an automated account replied, asking for more information about their situation. 
%
%
%
{Subsequently, \citet{cresci2023demystifying} confirmed that the data originated in the authors' previous work on detecting natural disasters using social media \cite{avvenuti2017hybrid}.} 

Other datasets yield classifiers which are similarly suggestive of their sampling and labeling procedures. 
The third tree in \Cref{fig:example-trees} shows that the account creation date is an important feature in distinguishing bots from humans in this dataset. 
This may result from the authors targeting the collection of \textit{spam} accounts on Twitter, for about eight months starting in December 2009. 
Active spam accounts may have high turnover on the platform if they are reported by users or the platform targets them for suspension, so very few spambots in the dataset were created more than a month before data collection began. 
The originators of the dataset make a similar observation \cite{lee2011a}. 
For \data{twibot-20}, the depth one decision tree reported in \Cref{table:sdt-summary} checks if the user is verified or not, and achieves nearly state-of-the-art performance on just this one feature. 
This may be an artifact of \textit{both} the sampling and labeling strategy. 
The authors collected accounts by starting from well-known (verified) seed users and collecting the network around those users using a breadth-first traversal \cite{feng2021twibot}, and we expect many verified users to follow each other. 
Accounts were labeled by crowdworkers and discarded if annotators did not sufficiently agree with each other, but verified accounts were automatically labeled human and so they were not at risk of being excluded. 
The other datasets we consider yield similar analyses.
Shallow decision trees reveal that simple models, suggestive of sampling and labeling heuristics, are powerful predictors.

The cases where our models underperform relative to the state of the art are informative:
the state-of-the-art model for \data{rtbust-2019} uses the time between a user's retweets, which we did not have access to.  
The bots in the dataset were identified by ``suspicious'' temporal retweeting patterns, so a simple decision rule with access to inter-tweet time may yield much higher performance.
%
\data{feedback-2019} is a small dataset collected from accounts reported to be misclassified by an earlier version of Botometer, 
and this is a complex sampling mechanism that may be hard to capture with our simple decision rules. 
For \data{yang-2013}, the F1 score is dragged down by low recall (the percentage of bots that were classified as bots), which may be a result of the classifier biasing predictions toward the human label since the dataset itself is more than 90\% human. 
When a classifier is trained on a balanced subsample of this data, the F1 score of a depth four decision tree is within 3\% of the state of the art.


\begin{table}[]
\caption{Performance of our shallow decision trees (SDT) versus state-of-the-art (SOTA) on benchmark datasets.
}
\label{table:sdt-summary}
\small
\begin{tabular}{@{}p{0.25\columnwidth}llllrr@{}} \toprule
Dataset & SDT & Depth & SOTA & SDT - SOTA \\ 
& Acc./F1/bal.\ acc.\ &   &  & Acc./F1 \\ \midrule
\data{twibot-2020}& 0.82/0.86/0.80  & 1 & \cite{feng2022heterogeneity-aware} & -0.05/-0.03 \\
\data{feedback-2019} & 0.80/0.55/0.69  & 3 & \cite{guo2022social} & -0.01/-0.15 \\
\data{rtbust-2019}  & 0.71/0.73/0.71  & 4 & \cite{mazza2019rtbust} & -0.22/-0.14 \\
\data{pan-2019}  & 0.92/0.91/0.92  & 2 & \cite{geng2021satar} & -0.03/-0.04 \\
\data{midterm-2018} & 0.97/0.98/0.95  & 1 & \cite{giorgi2021characterizing} & -0.01/\;\;--- \\
\data{stock-2018} & 0.80/0.83/0.80  & 3 & \;\;---\;\; & \;\;---\;\;\,/\;\;--- \\
\data{cresci-2017} & 0.98/0.98/0.97  & 1 & \cite{kudugunta2018deep} & -0.02/-0.02 \\
\data{gilani-2017} & 0.77/0.72/0.76  & 3 & \cite{gilani2020classification} & -0.09/-0.11 \\
\data{cresci-2015} & 0.98/0.98/0.98  & 3 & \cite{cresci2015fame} & -0.01/-0.01 \\
\data{yang-2013} & 0.96/0.71/0.79  & 4 & \cite{yang2013empirical} & -0.03/-0.19 \\
\data{caverlee-2011} & 0.91/0.91/0.90  & 2 & \cite{lee2011a} & -0.08/-0.07 \\

\end{tabular}
\end{table}

%
%

We stress that these results are not intended to suggest that simple decision rules make useful classifiers for bot detection, as they can be in other domains~\cite{rudin2019stop}; instead, they reveal that bot detection classifiers are limited by the simple sampling and labeling procedures used to construct datasets.
If we believed that bot detection is a simple, low-dimensional classification task, we might accept that simple classifiers suffice in this domain, but intuitively we do not expect this to be the case. In what follows, we support this intuition by considering how classifiers \textit{generalize} across datasets.

%
%
%

\subsection{Cross-dataset generalization.} \label{subsec:cross-dataset-generalization}

Heuristic data collection and labeling practices can be useful when the sample space is also easy to describe.
If the sample space is simple, we should expect that classifiers trained on one dataset perform well on others.
We present evidence here that this is not the case, by showing that classifiers that perform well on a given dataset typically do not significantly outperform random guessing when tested on each of the others, even when using more expressive models.
Similarly, we find that classifiers trained on all but one dataset and tested on the held-out dataset do not perform significantly better than random guessing in most cases.
From this, we conclude that the best predictors for separating human and bot users are not consistent across datasets.

\textit{Train on one, test on another.} 
For each dataset, we train a random forest using \texttt{scikit-learn}'s default parameter settings with 100 trees on each dataset and evaluate test performance on each of the others, using an 80\%-20\% train-test split. 
We restrict the feature sets for all training and testing data to the counts of account followers, following, number of tweets and lists each user is on since these features are common to nearly all datasets.
%
This allows us to compare each classifier on a consistent feature set. 
\data{yang-2013} and \data{caverlee-2011} do not contain information on Twitter lists, so we did not include that feature in classifiers trained or tested on those datasets.
When we instead used the pairwise intersection of the features available for each train-test combination, the results were qualitatively the same.
%

For this experiment, we use random forests rather than shallow decision trees to limit the extent to which the poor performance of the models can be attributed insufficient expressiveness, though we did find similar results when using shallow decision trees. 
Further, many papers in the literature use random forests to achieve state-of-the-art performance (see, e.g., \ \cite{yang2013empirical, gilani2017classification}).
Although it is possible that more expressive models, like neural networks, could achieve better cross-dataset performance than random forests for this experiment, we believe this to be implausible, since we can already capture much of the predictive signal in these datasets with simple decision rules.
The results are summarized in \cref{tab:tootoa-rf-bacc}, where each row corresponds to the dataset used for training and each column to the dataset used for testing. 
%
%
%
We see qualitatively similar results for accuracy and F1 scores, but omit those tables for brevity. 

In \Cref{tab:tootoa-rf-bacc}, the diagonal entries largely reproduce the experiment from \cref{subsec:individual-benchmarks} using random forests, showing the unsurprising fact that these can fit each dataset well. 
Because we use a restricted feature set, the performance of on-diagonal entries is sometimes lower in \cref{tab:tootoa-rf-bacc} than the performance reported in \cref{table:sdt-summary}.
Most of the entries off of the main diagonal show model performance no better than (weighted) random guessing or assigning all examples to a single class.
 
In a few cases, we see balanced accuracies significantly higher than 0.5.
This is the case for classifiers tested on \data{cresci-2017} and \data{cresci-2015} as well as to a lesser extent for \data{midterm-2018} and \data{yang-2013}.
These numbers may be explained by overlap between the datasets, as is the case between \data{cresci-2017} and \data{yang-2013} noted in \Cref{sec:datasets}, or similar collection and labeling strategies conducted by the same research group, such as is the case with \data{midterm-2019}, \data{cresci-2017} and \data{cresci-2015}. 
Notably, training on \data{yang-2013} yeilds poor performance when testing the classifier on other datasets, but other datasets yield classifiers that perform well on \data{yang-2013}, suggesting that it is a dataset that is in some sense easy to test on and not very useful for training classifiers.
In some cases, we see significantly worse than 0.5 balanced accuracy. 
This may arise from the distributions of users being very different across these datasets. 
A model trained on a dataset where human accounts have very low activity and bots have very high activity will perform very poorly when evaluated on a dataset where these patterns are reversed. 
The generally unimpressive performance of most of these classifiers suggests that the collections of accounts in each dataset, both bot and human, sit in different parts of the universe of possible accounts. 
%
\begin{table}[]
  \centering
  \caption{Balanced accuracy of random forests trained on the row-indexed dataset and tested on the column-indexed one. 
  %
  %
  }
  \label{tab:tootoa-rf-bacc}
  \begin{tikzpicture}[]
  \matrix[matrix of nodes,row sep=-\pgflinewidth, column sep=-.39em,
nodes={{rectangle}},
column 1/.style={{anchor=east}},]{
\data{\small{twibot-2020}} & |[fill={rgb,255:red,142;green,142;blue,215}, value=0.72]|&|[fill={rgb,255:red,244;green,244;blue,223}, value=0.52]|&|[fill={rgb,255:red,250;green,250;blue,223}, value=0.51]|&|[fill={rgb,255:red,240;green,240;blue,222}, value=0.53]|&|[fill={rgb,255:red,234;green,234;blue,222}, value=0.54]|&|[fill={rgb,255:red,244;green,244;blue,223}, value=0.52]|&|[fill={rgb,255:red,224;green,224;blue,221}, value=0.56]|&|[fill={rgb,255:red,254;green,254;blue,223}, value=0.5]|&|[fill={rgb,255:red,254;green,254;blue,223}, value=0.5]|&|[fill={rgb,255:red,252;green,250;blue,219}, value=0.49]| \\
\data{\small{feedback-2019}} & |[fill={rgb,255:red,240;green,240;blue,222}, value=0.53]|&|[fill={rgb,255:red,158;green,158;blue,216}, value=0.69]|&|[fill={rgb,255:red,244;green,244;blue,223}, value=0.52]|&|[fill={rgb,255:red,234;green,234;blue,222}, value=0.54]|&|[fill={rgb,255:red,105;green,105;blue,212}, value=0.79]|&|[fill={rgb,255:red,168;green,168;blue,217}, value=0.67]|&|[fill={rgb,255:red,208;green,208;blue,220}, value=0.59]|&|[fill={rgb,255:red,162;green,162;blue,217}, value=0.68]|&|[fill={rgb,255:red,112;green,112;blue,213}, value=0.78]|&|[fill={rgb,255:red,198;green,198;blue,219}, value=0.61]| \\
\data{\small{rtbust-2019}} & |[fill={rgb,255:red,254;green,254;blue,223}, value=0.5]|&|[fill={rgb,255:red,224;green,224;blue,221}, value=0.56]|&|[fill={rgb,255:red,142;green,142;blue,215}, value=0.72]|&|[fill={rgb,255:red,174;green,174;blue,217}, value=0.66]|&|[fill={rgb,255:red,254;green,254;blue,223}, value=0.5]|&|[fill={rgb,255:red,236;green,214;blue,187}, value=0.42]|&|[fill={rgb,255:red,230;green,230;blue,222}, value=0.55]|&|[fill={rgb,255:red,152;green,152;blue,216}, value=0.7]|&|[fill={rgb,255:red,219;green,178;blue,156}, value=0.35]|&|[fill={rgb,255:red,229;green,198;blue,173}, value=0.39]| \\
\data{\small{stock-2018}} & |[fill={rgb,255:red,240;green,240;blue,222}, value=0.53]|&|[fill={rgb,255:red,240;green,240;blue,222}, value=0.53]|&|[fill={rgb,255:red,214;green,214;blue,220}, value=0.58]|&|[fill={rgb,255:red,105;green,105;blue,212}, value=0.79]|&|[fill={rgb,255:red,178;green,178;blue,218}, value=0.65]|&|[fill={rgb,255:red,174;green,174;blue,217}, value=0.66]|&|[fill={rgb,255:red,250;green,250;blue,223}, value=0.51]|&|[fill={rgb,255:red,86;green,86;blue,211}, value=0.83]|&|[fill={rgb,255:red,254;green,254;blue,223}, value=0.5]|&|[fill={rgb,255:red,239;green,220;blue,193}, value=0.43]| \\
\data{\small{midterm-2018}} & |[fill={rgb,255:red,220;green,220;blue,221}, value=0.57]|&|[fill={rgb,255:red,230;green,230;blue,222}, value=0.55]|&|[fill={rgb,255:red,224;green,224;blue,221}, value=0.56]|&|[fill={rgb,255:red,204;green,204;blue,220}, value=0.6]|&|[fill={rgb,255:red,40;green,40;blue,207}, value=0.92]|&|[fill={rgb,255:red,80;green,80;blue,210}, value=0.84]|&|[fill={rgb,255:red,252;green,250;blue,219}, value=0.49]|&|[fill={rgb,255:red,92;green,92;blue,211}, value=0.82]|&|[fill={rgb,255:red,80;green,80;blue,210}, value=0.84]|&|[fill={rgb,255:red,198;green,198;blue,219}, value=0.61]| \\
\data{\small{cresci-2017}} & |[fill={rgb,255:red,208;green,208;blue,220}, value=0.59]|&|[fill={rgb,255:red,208;green,208;blue,220}, value=0.59]|&|[fill={rgb,255:red,240;green,240;blue,222}, value=0.53]|&|[fill={rgb,255:red,220;green,220;blue,221}, value=0.57]|&|[fill={rgb,255:red,60;green,60;blue,209}, value=0.88]|&|[fill={rgb,255:red,30;green,30;blue,207}, value=0.94]|&|[fill={rgb,255:red,252;green,250;blue,219}, value=0.49]|&|[fill={rgb,255:red,33;green,33;blue,207}, value=0.93]|&|[fill={rgb,255:red,194;green,194;blue,219}, value=0.62]|&|[fill={rgb,255:red,102;green,102;blue,212}, value=0.8]| \\
\data{\small{gilani-2017}} & |[fill={rgb,255:red,244;green,244;blue,223}, value=0.52]|&|[fill={rgb,255:red,214;green,214;blue,220}, value=0.58]|&|[fill={rgb,255:red,252;green,250;blue,219}, value=0.49]|&|[fill={rgb,255:red,230;green,230;blue,222}, value=0.55]|&|[fill={rgb,255:red,229;green,198;blue,173}, value=0.39]|&|[fill={rgb,255:red,248;green,240;blue,210}, value=0.47]|&|[fill={rgb,255:red,147;green,147;blue,216}, value=0.71]|&|[fill={rgb,255:red,254;green,254;blue,223}, value=0.5]|&|[fill={rgb,255:red,249;green,243;blue,214}, value=0.48]|&|[fill={rgb,255:red,231;green,204;blue,179}, value=0.4]| \\
\data{\small{cresci-2015}} & |[fill={rgb,255:red,234;green,234;blue,222}, value=0.54]|&|[fill={rgb,255:red,224;green,224;blue,221}, value=0.56]|&|[fill={rgb,255:red,254;green,254;blue,223}, value=0.5]|&|[fill={rgb,255:red,250;green,250;blue,223}, value=0.51]|&|[fill={rgb,255:red,184;green,184;blue,218}, value=0.64]|&|[fill={rgb,255:red,147;green,147;blue,216}, value=0.71]|&|[fill={rgb,255:red,254;green,254;blue,223}, value=0.5]|&|[fill={rgb,255:red,9;green,9;blue,205}, value=0.98]|&|[fill={rgb,255:red,96;green,96;blue,212}, value=0.81]|&|[fill={rgb,255:red,158;green,158;blue,216}, value=0.69]| \\
\data{\small{yang-2013}} & |[fill={rgb,255:red,250;green,250;blue,223}, value=0.51]|&|[fill={rgb,255:red,240;green,240;blue,222}, value=0.53]|&|[fill={rgb,255:red,254;green,254;blue,223}, value=0.5]|&|[fill={rgb,255:red,249;green,243;blue,214}, value=0.48]|&|[fill={rgb,255:red,240;green,240;blue,222}, value=0.53]|&|[fill={rgb,255:red,204;green,204;blue,220}, value=0.6]|&|[fill={rgb,255:red,254;green,254;blue,223}, value=0.5]|&|[fill={rgb,255:red,208;green,208;blue,220}, value=0.59]|&|[fill={rgb,255:red,96;green,96;blue,212}, value=0.81]|&|[fill={rgb,255:red,214;green,214;blue,220}, value=0.58]| \\
\data{\small{caverlee-2011}} & |[fill={rgb,255:red,198;green,198;blue,219}, value=0.61]|&|[fill={rgb,255:red,220;green,220;blue,221}, value=0.57]|&|[fill={rgb,255:red,236;green,214;blue,187}, value=0.42]|&|[fill={rgb,255:red,248;green,240;blue,210}, value=0.47]|&|[fill={rgb,255:red,248;green,240;blue,210}, value=0.47]|&|[fill={rgb,255:red,102;green,102;blue,212}, value=0.8]|&|[fill={rgb,255:red,240;green,240;blue,222}, value=0.53]|&|[fill={rgb,255:red,92;green,92;blue,211}, value=0.82]|&|[fill={rgb,255:red,230;green,230;blue,222}, value=0.55]|&|[fill={rgb,255:red,56;green,56;blue,209}, value=0.89]| \\
};
\node[label={[label distance=0.5cm,text depth=-1ex,rotate=45]left: \data{\small{twibot-2020}} }] at (-1.1,-2.6) {};
\node[label={[label distance=0.5cm,text depth=-1ex,rotate=45]left: \data{\small{feedback-2019}} }] at (-0.5000000000000001,-2.6) {};
\node[label={[label distance=0.5cm,text depth=-1ex,rotate=45]left: \data{\small{rtbust-2019}} }] at (0.09999999999999987,-2.6) {};
\node[label={[label distance=0.5cm,text depth=-1ex,rotate=45]left: \data{\small{stock-2018}} }] at (0.6999999999999997,-2.6) {};
\node[label={[label distance=0.5cm,text depth=-1ex,rotate=45]left: \data{\small{midterm-2018}} }] at (1.2999999999999998,-2.6) {};
\node[label={[label distance=0.5cm,text depth=-1ex,rotate=45]left: \data{\small{cresci-2017}} }] at (1.9,-2.6) {};
\node[label={[label distance=0.5cm,text depth=-1ex,rotate=45]left: \data{\small{gilani-2017}} }] at (2.4999999999999996,-2.6) {};
\node[label={[label distance=0.5cm,text depth=-1ex,rotate=45]left: \data{\small{cresci-2015}} }] at (3.1,-2.6) {};
\node[label={[label distance=0.5cm,text depth=-1ex,rotate=45]left: \data{\small{yang-2013}} }] at (3.6999999999999997,-2.6) {};
\node[label={[label distance=0.5cm,text depth=-1ex,rotate=45]left: \data{\small{caverlee-2011}} }] at (4.299999999999999,-2.6) {};
\end{tikzpicture}
\end{table}
%
%


%

\textit{Leave-one-dataset-out.} 
Following \cite{echeverria2018lobo, sayyadiharikandeh2020detection}, we also train random forest classifiers on all but one dataset and test on the held-out dataset. 
We report the results in \cref{table:loo}.
In most cases, the signal contained in all but one dataset is not sufficient to predict the labels on the held out dataset. 
For the cases where this is not true, the datasets share some data in common or were collected and labeled using similar strategies.
These results suggest that the best predictors in each dataset are different, and therefore that these benchmark datasets, even when combined, do not generalize to perform well on the others. 
%
%
In the rest of this section, we explore generalization across datasets further, examining whether making assumptions about the types of bots that exist in each dataset can be used to build generalizable predictors.

\begin{table}[]
    \caption{Leave-one-dataset-out on benchmark datasets.}
    \small
    \begin{tabular}{@{}lll@{}} \toprule
    & In-sample & Out-of-sample\\
    Dataset left out & Acc./F1/bal.\ acc.\ & Acc./F1/bal.\ acc.\ \\ \midrule
\data{twibot-2020} & 0.80/0.84/0.78 & 0.52/0.44/0.55 \\
\data{feedback-2019} & 0.78/0.82/0.77 & 0.64/0.37/0.56 \\
\data{rtbust-2019} & 0.78/0.82/0.76 & 0.52/0.31/0.53 \\
\data{midterm-2018} & 0.78/0.82/0.77 & 0.77/0.85/0.77 \\
\data{stock-2018} & 0.76/0.81/0.75 & 0.55/0.49/0.56 \\
\data{cresci-2017} & 0.77/0.82/0.75 & 0.83/0.88/0.84 \\
\data{gilani-2017} & 0.79/0.84/0.78 & 0.58/0.22/0.52 \\
\data{cresci-2015} & 0.81/0.86/0.78 & 0.87/0.90/0.83 \\
\data{yang-2013} & 0.84/0.88/0.83 & 0.32/0.21/0.62 \\
\data{caverlee-2011} & 0.71/0.71/0.71 & 0.56/0.56/0.57 \\

    \end{tabular}
    \label{table:loo}
\end{table}

\subsection{Generalization using a bot taxonomy.} \label{subsec:bot-types}

We observe in the previous analysis that classifiers trained on one or more benchmark datasets do not give good performance on the others. 
However, previous work \cite{sayyadiharikandeh2020detection, dimitriadis2021social} has argued that different datasets may contain different types of bots---e.g., simple bots, spammers, fake followers, self-declared, political bots, and other bots (a catch-all category)---which could account for poor out-of-sample performance in cases where the bot types are different.
The need to combine data from different populations is common in prediction problems. 
%
For example, to make predictions on patient data from adult and pediatric hospitals, it is necessary to combine data across age groups in ways that account for population differences.
Similarly, bot detection may require differentiating between and combining data from qualitatively different types of bots.
%
%

Following Botometer \cite{sayyadiharikandeh2020detection}, we  define combined bot-type datasets as follows. \data{simple} consists of the bot accounts from \data{caverlee\hyp{}2011}; \data{spammers} includes the social or traditional spambots datasets and \data{pronbots\hyp{}2019}; \data{fake\hyp{}followers} consists of \data{fake\hyp{}followers\hyp{}2015} and \data{vendor\hyp{}purchased\hyp{}2019}; \data{other\hyp{}bots} consists of \data{feedback\hyp{}2019} and \data{rtbust\hyp{}2019}; \data{financial\hyp{}bots} consists of \data{stock\hyp{}2018}. 
We do not have access to data for the roughly 500 accounts in \data{astroturf-2020}, included in \data{political\hyp{}bots} in \cite{sayyadiharikandeh2020detection}. 
%

In \cref{tab:bot-types-performance}, we show that these datasets admit accurate yet simple classifiers for detecting a bot type against a sample of human accounts, indicating as in \cref{subsec:individual-benchmarks} that the predictive signal contained in each bot type-human dataset is also simple.
For each type of bot, we take equal-sized random samples of human accounts and accounts from the bot type and learn a shallow decision tree. 
We report accuracy and F1, along with the depth that achieves the stated test performance for an 80\%-20\% train/test split. 
Since we balance the datasets, accuracy and balanced accuracy are equivalent and we do not report the latter.
As before, we favor shallower decision trees when performance is comparable across depths.

Across the bot types, the simple decision rules achieve high accuracy against a baseline of 0.5 (random guessing or a naive classifier) on this task, summarized in \Cref{tab:bot-types-performance}. 
%
%
The performance on \data{political\hyp{}bots} is perfect, perhaps as a result of the small size of this bot type in our data.
We see only modest performance improvements by using a random forest rather than a simple decision rule and we omit the details of those results here. 

%


\begin{table}[]

    \caption{Distinguishing each bot type from humans. }
    \small
    \begin{tabular}{@{}lll@{}} \toprule
    
    Bot type & Acc/F1 & Depth\\ \midrule
\data{simple} & 0.86/0.86  & 3 \\
\data{spammers} & 0.89/0.89  & 3 \\
\data{fake-followers} & 0.94/0.94  & 4 \\
\data{self-declared} & 0.88/0.87  & 3 \\
\data{political\hyp{}bots} & 1.00/1.00  & 3 \\
\data{other-bots} & 0.76/0.74  & 2 \\
\data{financial-bots} & 0.73/0.75  & 4 \\

    \end{tabular}
    \label{tab:bot-types-performance}
\end{table}

We next learn classifiers that predict which dataset an account comes from within bot types in order to understand whether accounts from different datasets in the same category are substantively similar to one another.

\subsection{Distinguishing datasets within bot types.} \label{subsec:distinguishing-bot-types}

Continuing with the taxonomic viewpoint, here we address whether these datasets can be used to train models that generalize well \textit{only within a single class of bots}. We provide evidence towards the negative by showing that it is possible to distinguish accounts from different datasets within a bot type. 
Even when restricted to the subspace of spam bots, for instance, the constituent datasets in that class are sampled in such distinct ways that a classifier can easily distinguish them. We also observe a similar phenomenon with the human accounts, indicating that the strategies used to gather and label human accounts also results in samples being drawn from different parts of that sample space. 
In \cref{tab:distinguishing-bot-types}, we report accuracy and balanced accuracy for the task of predicting which dataset an account comes from for accounts within a given type.
Rather than a binary prediction task of bot or not, this is a multiclass prediction task where each account is labeled with the dataset it comes from, omiting the classes that consist of only one dataset. 
We train shallow decision trees to predict these labels,  summarized in \cref{tab:distinguishing-bot-types}. Again we see minor performance improvements by using random forest classifiers, but prefer the decision trees for their interpretability.
%

%

For \data{humans}, we have access to six datasets, so a naive or random classifier should achieve a balanced accuracy of less than 0.17. 
However, we achieve accuracy greater than 0.65 and balanced accuracy of greater than 0.4, indicating that this classifier can identify which dataset a human account belongs to with probability significantly better than random chance.
For \data{spammers}, we achieve balanced accuracy of 0.75 and nearly perfect accuracy on seven datasets, against a naive baseline of less than 0.15. 
The classifier here gains predictive power from artifacts of the sampling strategies used to collect the data. As examples, most of the accounts in \data{pronbots-2019} liked several tweets whereas other spammers did not; accounts in \data{social-spambots-1} sent many tweets in Italian whereas other spammers used English or only tweeted URLs; still other datasets can be separated by basic account activity features like the number of accounts followed.

\begin{table}[]
    \caption{Distinguishing an account's dataset by type. }

    \small
    \begin{tabular}{llll}
    \toprule
        Type & Acc./bal.\ acc.\ & Depth & Num. datasets \\
        \midrule
\data{humans} & 0.67/0.43 & 3 & 6 \\
\data{spammers} & 0.97/0.75 & 4 & 7 \\
\data{fake followers} & 0.97/0.94 & 1 & 2 \\
\data{other bots} & 0.91/0.77 & 2 & 3 \\
    \end{tabular}
        \label{tab:distinguishing-bot-types}
\end{table}

\section{Conclusion \& Discussion}
\label{sec:discussion}

In this paper, we provided evidence that Twitter bot detection datasets are limited by their simple collection strategies, explaining their poor generalization.
%
These findings suggest that a limiting factor in advancing bot detection research is a lack of availability of robust, high quality data.
%
If accounts are improperly discarded for being likely bots because they have not liked enough tweets or their account was created at the wrong time or they were accessed from a particular kind of device, that introduces errors into a downstream analysis and if any of those features correlate with the topic of interest, these errors may bias the conclusions of that analysis.

This work also highlights the broader issue of how opaque machine learning techniques may obscure certain flaws in the underlying data.{ While we examine the Twitter bot detection ecosystem in this work, our methods and recommendations should apply broadly to the study of any online social media platform and to applied machine learning research wherever datasets are reused across contexts.}  We encourage researchers who originate and publish datasets to be explicit about their sampling and labeling procedures, perhaps using existing documentation tools~\cite{gebru2021datasheets}.
Researchers and engineers building bot detection tools should carefully examine their training data, possibly via the use of simple models like shallow decision trees, to ensure the data captures the complexity of the space the more expressive model attempts to describe.
Finally, those using bot detection as a preprocessing step should carefully consider how errors might propagate through this process and what kinds of biases might be present in their analysis as a result. 
Given how seriously the academic community and general public treat the problem of bot detection on social media, we also hope that the platforms themselves can facilitate work in this area by providing rich and robust data with high quality ground truth labels.
%

%


\pagebreak

\bibliographystyle{ACM-Reference-Format}
\bibliography{botdetection, methods}


\begin{thebibliography}{77}


\ifx \showCODEN    \undefined \def \showCODEN     #1{\unskip}     \fi
\ifx \showDOI      \undefined \def \showDOI       #1{#1}\fi
\ifx \showISBNx    \undefined \def \showISBNx     #1{\unskip}     \fi
\ifx \showISBNxiii \undefined \def \showISBNxiii  #1{\unskip}     \fi
\ifx \showISSN     \undefined \def \showISSN      #1{\unskip}     \fi
\ifx \showLCCN     \undefined \def \showLCCN      #1{\unskip}     \fi
\ifx \shownote     \undefined \def \shownote      #1{#1}          \fi
\ifx \showarticletitle \undefined \def \showarticletitle #1{#1}   \fi
\ifx \showURL      \undefined \def \showURL       {\relax}        \fi
\providecommand\bibfield[2]{#2}
\providecommand\bibinfo[2]{#2}
\providecommand\natexlab[1]{#1}
\providecommand\showeprint[2][]{arXiv:#2}

\bibitem[Alothali et~al\mbox{.}(2022)]%
        {alothali2022bot-mgat:}
\bibfield{author}{\bibinfo{person}{Eiman Alothali}, \bibinfo{person}{Motamen Salih}, \bibinfo{person}{Kadhim Hayawi}, {and} \bibinfo{person}{Hany Alashwal}.} \bibinfo{year}{2022}\natexlab{}.
\newblock \showarticletitle{Bot-MGAT: A Transfer Learning Model Based on a Multi-View Graph Attention Network to Detect Social Bots}.
\newblock \bibinfo{journal}{\emph{Applied Sciences}}  \bibinfo{volume}{16} (\bibinfo{year}{2022}).
\newblock
\urldef\tempurl%
\url{https://doi.org/10.3390/app12168117}
\showDOI{\tempurl}


\bibitem[Avvenuti et~al\mbox{.}(2017)]%
        {avvenuti2017hybrid}
\bibfield{author}{\bibinfo{person}{Marco Avvenuti}, \bibinfo{person}{Salvatore Bellomo}, \bibinfo{person}{Stefano Cresci}, \bibinfo{person}{Mariantonietta~Noemi La~Polla}, {and} \bibinfo{person}{Maurizio Tesconi}.} \bibinfo{year}{2017}\natexlab{}.
\newblock \showarticletitle{Hybrid crowdsensing: A novel paradigm to combine the strengths of opportunistic and participatory crowdsensing}. In \bibinfo{booktitle}{\emph{Proceedings of the 26th international conference on World Wide Web companion}}. \bibinfo{pages}{1413--1421}.
\newblock


\bibitem[Badawy et~al\mbox{.}(2018)]%
        {badawy2018analyzing}
\bibfield{author}{\bibinfo{person}{Adam Badawy}, \bibinfo{person}{Emilio Ferrara}, {and} \bibinfo{person}{Kristina Lerman}.} \bibinfo{year}{2018}\natexlab{}.
\newblock \showarticletitle{Analyzing the Digital Traces of Political Manipulation: The 2016 Russian Interference Twitter Campaign}. In \bibinfo{booktitle}{\emph{2018 IEEE/ACM International Conference on Advances in Social Networks Analysis and Mining (ASONAM)}}. \bibinfo{pages}{258--265}.
\newblock
\urldef\tempurl%
\url{https://doi.org/10.1109/ASONAM.2018.8508646}
\showDOI{\tempurl}


\bibitem[Bastos and Mercea(2019)]%
        {bastos2019brexit}
\bibfield{author}{\bibinfo{person}{Marco~T. Bastos} {and} \bibinfo{person}{Dan Mercea}.} \bibinfo{year}{2019}\natexlab{}.
\newblock \showarticletitle{The Brexit Botnet and User-Generated Hyperpartisan News}.
\newblock \bibinfo{journal}{\emph{Social Science Computer Review}} \bibinfo{volume}{37}, \bibinfo{number}{1} (\bibinfo{year}{2019}), \bibinfo{pages}{38--54}.
\newblock
\urldef\tempurl%
\url{https://doi.org/10.1177/0894439317734157}
\showDOI{\tempurl}
\showeprint{https://doi.org/10.1177/0894439317734157}


\bibitem[Bessi and Ferrara(2016)]%
        {bessi2016social}
\bibfield{author}{\bibinfo{person}{Alessandro Bessi} {and} \bibinfo{person}{Emilio Ferrara}.} \bibinfo{year}{2016}\natexlab{}.
\newblock \showarticletitle{Social bots distort the 2016 US Presidential election online discussion.}
\newblock \bibinfo{journal}{\emph{First Monday}} \bibinfo{volume}{21}, \bibinfo{number}{11} (\bibinfo{year}{2016}).
\newblock
\urldef\tempurl%
\url{https://doi.org/10.5210/fm.v21i11.7090}
\showDOI{\tempurl}


\bibitem[Bouzy(2018)]%
        {botsent_2022}
\bibfield{author}{\bibinfo{person}{Christopher Bouzy}.} \bibinfo{year}{2018}\natexlab{}.
\newblock \bibinfo{title}{Bot sentinel, Platform developed to detect and track political bots, trollbots, and untrustworthy accounts.}
\newblock
\newblock
\urldef\tempurl%
\url{https://botsentinel.com/}
\showURL{%
\tempurl}


\bibitem[Broniatowski et~al\mbox{.}(2018)]%
        {broniatowski2018}
\bibfield{author}{\bibinfo{person}{David~A. Broniatowski}, \bibinfo{person}{Amelia~M. Jamison}, \bibinfo{person}{SiHua Qi}, \bibinfo{person}{Lulwah AlKulaib}, \bibinfo{person}{Tao Chen}, \bibinfo{person}{Adrian Benton}, \bibinfo{person}{Sandra~C. Quinn}, {and} \bibinfo{person}{Mark Dredze}.} \bibinfo{year}{2018}\natexlab{}.
\newblock \showarticletitle{Weaponized Health Communication: Twitter Bots and Russian Trolls Amplify the Vaccine Debate}.
\newblock \bibinfo{journal}{\emph{American Journal of Public Health}} \bibinfo{volume}{108}, \bibinfo{number}{10} (\bibinfo{year}{2018}), \bibinfo{pages}{1378--1384}.
\newblock
\urldef\tempurl%
\url{https://doi.org/10.2105/AJPH.2018.304567}
\showDOI{\tempurl}
\showeprint{https://doi.org/10.2105/AJPH.2018.304567}
\newblock
\shownote{PMID: 30138075}.


\bibitem[Choi et~al\mbox{.}(2020)]%
        {choi2020rumor}
\bibfield{author}{\bibinfo{person}{Daejin Choi}, \bibinfo{person}{Selin Chun}, \bibinfo{person}{Hyunchul Oh}, \bibinfo{person}{Jinyoung Han}, {and} \bibinfo{person}{Ted Kwon}.} \bibinfo{year}{2020}\natexlab{}.
\newblock \showarticletitle{Rumor Propagation is Amplified by Echo Chambers in Social Media}.
\newblock \bibinfo{journal}{\emph{Scientific Reports}} \bibinfo{volume}{10}, \bibinfo{number}{310} (\bibinfo{year}{2020}).
\newblock
\urldef\tempurl%
\url{https://doi.org/10.1038/s41598-019-57272-3}
\showDOI{\tempurl}


\bibitem[Chu et~al\mbox{.}(2012)]%
        {chu2012detecting}
\bibfield{author}{\bibinfo{person}{Zi Chu}, \bibinfo{person}{Steven Gianvecchio}, \bibinfo{person}{Haining Wang}, {and} \bibinfo{person}{Sushil Jajodia}.} \bibinfo{year}{2012}\natexlab{}.
\newblock \showarticletitle{Detecting Automation of Twitter Accounts: Are You a Human, Bot, or Cyborg?}
\newblock \bibinfo{journal}{\emph{IEEE Transactions on Dependable and Secure Computing}} \bibinfo{volume}{9}, \bibinfo{number}{6} (\bibinfo{year}{2012}), \bibinfo{pages}{811--824}.
\newblock
\urldef\tempurl%
\url{https://doi.org/10.1109/TDSC.2012.75}
\showDOI{\tempurl}


\bibitem[Cresci(2020)]%
        {cresci2020decade}
\bibfield{author}{\bibinfo{person}{Stefano Cresci}.} \bibinfo{year}{2020}\natexlab{}.
\newblock \showarticletitle{A Decade of Social Bot Detection}.
\newblock \bibinfo{journal}{\emph{Commun. ACM}} \bibinfo{volume}{63}, \bibinfo{number}{10} (\bibinfo{date}{sep} \bibinfo{year}{2020}), \bibinfo{pages}{72–83}.
\newblock
\showISSN{0001-0782}
\urldef\tempurl%
\url{https://doi.org/10.1145/3409116}
\showDOI{\tempurl}


\bibitem[Cresci et~al\mbox{.}(2017a)]%
        {cresci2017the}
\bibfield{author}{\bibinfo{person}{Stefano Cresci}, \bibinfo{person}{Roberto Di~Pietro}, \bibinfo{person}{Marinella Petrocchi}, \bibinfo{person}{Angelo Spognardi}, {and} \bibinfo{person}{Maurizio Tesconi}.} \bibinfo{year}{2017}\natexlab{a}.
\newblock \showarticletitle{The paradigm-shift of social spambots: Evidence, theories, and tools for the arms race.}. In \bibinfo{booktitle}{\emph{26th International Conference on World Wide Web}}. \bibinfo{publisher}{International World Wide Web Conferences Steering Committee}, \bibinfo{pages}{963--972}.
\newblock
\urldef\tempurl%
\url{https://doi.org/10.1145/3041021.3055135}
\showDOI{\tempurl}


\bibitem[Cresci et~al\mbox{.}(2023)]%
        {cresci2023demystifying}
\bibfield{author}{\bibinfo{person}{Stefano Cresci}, \bibinfo{person}{Roberto Di~Pietro}, \bibinfo{person}{Angelo Spognardi}, \bibinfo{person}{Maurizio Tesconi}, {and} \bibinfo{person}{Marinella Petrocchi}.} \bibinfo{year}{2023}\natexlab{}.
\newblock \showarticletitle{Demystifying Misconceptions in Social Bots Research}.
\newblock \bibinfo{journal}{\emph{arXiv preprint arXiv:2303.17251}} (\bibinfo{year}{2023}).
\newblock


\bibitem[Cresci et~al\mbox{.}(2019)]%
        {cresci2019fake}
\bibfield{author}{\bibinfo{person}{Stefano Cresci}, \bibinfo{person}{Fabrizio Lillo}, \bibinfo{person}{Daniele Regoli}, \bibinfo{person}{Serena Tardelli}, {and} \bibinfo{person}{Maurizio Tesconi}.} \bibinfo{year}{2019}\natexlab{}.
\newblock \showarticletitle{\$ FAKE: Evidence of Spam and Bot Activity in Stock Microblogs on Twitter.}
\newblock  (\bibinfo{year}{2019}).
\newblock


\bibitem[Cresci et~al\mbox{.}(2015a)]%
        {cresci2015fame}
\bibfield{author}{\bibinfo{person}{Stefano Cresci}, \bibinfo{person}{Roberto~Di Pietro}, \bibinfo{person}{Marinella Petrocchi}, \bibinfo{person}{Angelo Spognardi}, {and} \bibinfo{person}{Maurizio Tesconi}.} \bibinfo{year}{2015}\natexlab{a}.
\newblock \showarticletitle{Fame for sale: Efficient detection of fake Twitter followers}.
\newblock \bibinfo{journal}{\emph{Decision Support Systems}}  \bibinfo{volume}{80} (\bibinfo{year}{2015}), \bibinfo{pages}{56--71}.
\newblock
\urldef\tempurl%
\url{https://doi.org/10.1016/j.dss.2015.09.003}
\showDOI{\tempurl}


\bibitem[Cresci et~al\mbox{.}(2016)]%
        {cresci2016dna-inspired}
\bibfield{author}{\bibinfo{person}{Stefano Cresci}, \bibinfo{person}{Roberto~Di Pietro}, \bibinfo{person}{Marinella Petrocchi}, \bibinfo{person}{Angelo Spognardi}, {and} \bibinfo{person}{Maurizio Tesconi}.} \bibinfo{year}{2016}\natexlab{}.
\newblock \showarticletitle{DNA-Inspired Online Behavioral Modeling and Its Application to Spambot Detection}.
\newblock \bibinfo{journal}{\emph{IEEE Intelligent Systems}} \bibinfo{volume}{31}, \bibinfo{number}{5} (\bibinfo{year}{2016}), \bibinfo{pages}{58--64}.
\newblock
\urldef\tempurl%
\url{https://doi.org/10.1109/MIS.2016.29}
\showDOI{\tempurl}


\bibitem[Cresci et~al\mbox{.}(2017b)]%
        {cresci2017exploiting}
\bibfield{author}{\bibinfo{person}{Stefano Cresci}, \bibinfo{person}{Roberto~Di Pietro}, \bibinfo{person}{Marinella Petrocchi}, \bibinfo{person}{Angelo Spognardi}, {and} \bibinfo{person}{Maurizio Tesconi}.} \bibinfo{year}{2017}\natexlab{b}.
\newblock \showarticletitle{Exploiting digital DNA for the analysis of similarities in Twitter behaviour}. In \bibinfo{booktitle}{\emph{IEEE International Conference on Data Science and Advanced Analytics (DSAA)}}. \bibinfo{pages}{686--695}.
\newblock
\urldef\tempurl%
\url{https://doi.org/10.1109/DSAA.2017.57}
\showDOI{\tempurl}


\bibitem[Cresci et~al\mbox{.}(2018)]%
        {cresci2018social}
\bibfield{author}{\bibinfo{person}{Stefano Cresci}, \bibinfo{person}{Roberto~Di Pietro}, \bibinfo{person}{Marinella Petrocchi}, \bibinfo{person}{Angelo Spognardi}, {and} \bibinfo{person}{Maurizio Tesconi}.} \bibinfo{year}{2018}\natexlab{}.
\newblock \showarticletitle{Social Fingerprinting: Detection of Spambot Groups Through DNA-Inspired Behavioral Modeling}.
\newblock \bibinfo{journal}{\emph{IEEE Transactions on Dependable and Secure Computing}} \bibinfo{volume}{15}, \bibinfo{number}{4} (\bibinfo{year}{2018}), \bibinfo{pages}{561--576}.
\newblock
\urldef\tempurl%
\url{https://doi.org/10.1109/TDSC.2017.2681672}
\showDOI{\tempurl}


\bibitem[Cresci et~al\mbox{.}(2015b)]%
        {cresci2015linguistically}
\bibfield{author}{\bibinfo{person}{Stefano Cresci}, \bibinfo{person}{Maurizio Tesconi}, \bibinfo{person}{Andrea Cimino}, {and} \bibinfo{person}{Felice Dell'Orletta}.} \bibinfo{year}{2015}\natexlab{b}.
\newblock \showarticletitle{A Linguistically-Driven Approach to Cross-Event Damage Assessment of Natural Disasters from Social Media Messages}. In \bibinfo{booktitle}{\emph{Proceedings of the 24th International Conference on World Wide Web}} (Florence, Italy) \emph{(\bibinfo{series}{WWW '15 Companion})}. \bibinfo{publisher}{Association for Computing Machinery}, \bibinfo{address}{New York, NY, USA}, \bibinfo{pages}{1195–1200}.
\newblock
\showISBNx{9781450334730}
\urldef\tempurl%
\url{https://doi.org/10.1145/2740908.2741722}
\showDOI{\tempurl}


\bibitem[Dimitriadis et~al\mbox{.}(2021)]%
        {dimitriadis2021social}
\bibfield{author}{\bibinfo{person}{Ilias Dimitriadis}, \bibinfo{person}{Konstantinos Georgiou}, {and} \bibinfo{person}{Athena Vakali}.} \bibinfo{year}{2021}\natexlab{}.
\newblock \showarticletitle{Social Botomics: A Systematic Ensemble ML Approach for Explainable and Multi-Class Bot Detection}.
\newblock \bibinfo{journal}{\emph{Applied Sciences}} \bibinfo{volume}{11}, \bibinfo{number}{21} (\bibinfo{year}{2021}).
\newblock
\urldef\tempurl%
\url{https://doi.org/10.3390/app11219857}
\showDOI{\tempurl}


\bibitem[Echeverria et~al\mbox{.}(2018)]%
        {echeverria2018lobo}
\bibfield{author}{\bibinfo{person}{Juan Echeverria}, \bibinfo{person}{Emiliano De~Cristofaro}, \bibinfo{person}{Nicolas Kourtellis}, \bibinfo{person}{Ilias Leontiadis}, \bibinfo{person}{Gianluca Stringhini}, {and} \bibinfo{person}{Shi Zhou}.} \bibinfo{year}{2018}\natexlab{}.
\newblock \showarticletitle{LOBO: Evaluation of Generalization Deficiencies in Twitter Bot Classifiers}. In \bibinfo{booktitle}{\emph{Proceedings of the 34th Annual Computer Security Applications Conference}} (San Juan, PR, USA) \emph{(\bibinfo{series}{ACSAC '18})}. \bibinfo{publisher}{Association for Computing Machinery}, \bibinfo{address}{New York, NY, USA}, \bibinfo{pages}{137–146}.
\newblock
\showISBNx{9781450365697}
\urldef\tempurl%
\url{https://doi.org/10.1145/3274694.3274738}
\showDOI{\tempurl}


\bibitem[Elmas et~al\mbox{.}(2022)]%
        {elmas2022characterizing}
\bibfield{author}{\bibinfo{person}{Tu{\u{g}}rulcan Elmas}, \bibinfo{person}{Rebekah Overdorf}, {and} \bibinfo{person}{Karl Aberer}.} \bibinfo{year}{2022}\natexlab{}.
\newblock \showarticletitle{Characterizing Retweet Bots: The Case of Black Market Accounts}. In \bibinfo{booktitle}{\emph{Proceedings of the International AAAI Conference on Web and Social Media}}, Vol.~\bibinfo{volume}{16}. \bibinfo{pages}{171--182}.
\newblock


\bibitem[Feng et~al\mbox{.}(2022)]%
        {feng2022heterogeneity-aware}
\bibfield{author}{\bibinfo{person}{Shangbin Feng}, \bibinfo{person}{Zhaoxuan Tan}, {and} \bibinfo{person}{Minnan~Luo Rui L~and}.} \bibinfo{year}{2022}\natexlab{}.
\newblock \showarticletitle{Heterogeneity-aware Twitter Bot Detection with Relational Graph Transformers}. In \bibinfo{booktitle}{\emph{AAAI Conference on Artificial Intelligence}}. \bibinfo{publisher}{Association for the Advancement of Artificial Intelligence}, \bibinfo{pages}{3977--3985}.
\newblock
\urldef\tempurl%
\url{https://doi.org/10.1609/aaai.v36i4.20314}
\showDOI{\tempurl}


\bibitem[Feng et~al\mbox{.}(2021b)]%
        {geng2021satar}
\bibfield{author}{\bibinfo{person}{Shangbin Feng}, \bibinfo{person}{Herun Wan}, \bibinfo{person}{Ningnan Wang}, \bibinfo{person}{Jundong Li}, {and} \bibinfo{person}{Minnan Luo}.} \bibinfo{year}{2021}\natexlab{b}.
\newblock \showarticletitle{SATAR: A Self-supervised Approach to Twitter Account Representation Learning and its Application in Bot Detection}. In \bibinfo{booktitle}{\emph{30th ACM International Conference on Information \& Knowledge Management (CIKM)}}. \bibinfo{publisher}{Association for Computing Machinery}, \bibinfo{pages}{3808--3817}.
\newblock
\urldef\tempurl%
\url{https://doi.org/10.1145/3459637.3481949}
\showDOI{\tempurl}


\bibitem[Feng et~al\mbox{.}(2021c)]%
        {feng2021twibot}
\bibfield{author}{\bibinfo{person}{Shangbin Feng}, \bibinfo{person}{Herun Wan}, \bibinfo{person}{Ningnan Wang}, \bibinfo{person}{Jundong Li}, {and} \bibinfo{person}{Minnan Luo}.} \bibinfo{year}{2021}\natexlab{c}.
\newblock \showarticletitle{TwiBot-20: A Comprehensive Twitter Bot Detection Benchmark}.
\newblock  (\bibinfo{year}{2021}), \bibinfo{pages}{4485--4494}.
\newblock


\bibitem[Feng et~al\mbox{.}(2021a)]%
        {feng2021botrgcn}
\bibfield{author}{\bibinfo{person}{Shangbin Feng}, \bibinfo{person}{Herun Wan}, \bibinfo{person}{Ningnan Wang}, {and} \bibinfo{person}{Minnan Luo}.} \bibinfo{year}{2021}\natexlab{a}.
\newblock \showarticletitle{BotRGCN: Twitter Bot Detection with Relational Graph Convolutional Networks}. In \bibinfo{booktitle}{\emph{2021 IEEE/ACM International Conference on Advances in Social Networks Analysis and Mining}}. \bibinfo{publisher}{Association for Computing Machinery}, \bibinfo{pages}{236--239}.
\newblock
\urldef\tempurl%
\url{https://doi.org/10.1145/3487351.3488336}
\showDOI{\tempurl}


\bibitem[Ferrara(2017)]%
        {ferrara2017disinformation}
\bibfield{author}{\bibinfo{person}{Emilio Ferrara}.} \bibinfo{year}{2017}\natexlab{}.
\newblock \showarticletitle{Disinformation and Social Bot Operations in the Run Up to the 2017 French Presidential Election}.
\newblock \bibinfo{journal}{\emph{First Monday}} \bibinfo{volume}{22}, \bibinfo{number}{8} (\bibinfo{year}{2017}).
\newblock
\urldef\tempurl%
\url{https://doi.org/10.5210/fm.v22i18.8005}
\showDOI{\tempurl}


\bibitem[Ferrara(2020)]%
        {ferrara2020covid}
\bibfield{author}{\bibinfo{person}{Emilio Ferrara}.} \bibinfo{year}{2020}\natexlab{}.
\newblock \showarticletitle{What types of {COVID}-19 conspiracies are populated by Twitter bots?}
\newblock \bibinfo{journal}{\emph{First Monday}} (\bibinfo{date}{05} \bibinfo{year}{2020}).
\newblock
\urldef\tempurl%
\url{https://doi.org/10.5210/fm.v25i6.10633}
\showDOI{\tempurl}


\bibitem[Gallwitz and Kreil(2021)]%
        {gallwitz2021therise}
\bibfield{author}{\bibinfo{person}{Florian Gallwitz} {and} \bibinfo{person}{Michael Kreil}.} \bibinfo{year}{2021}\natexlab{}.
\newblock \bibinfo{title}{The Rise and Fall of 'Social Bot' Research}.
\newblock
\newblock
\urldef\tempurl%
\url{https://ssrn.com/abstract=3814191}
\showURL{%
\tempurl}


\bibitem[Gallwitz and Kreil(2022)]%
        {gallwitz2022investigating}
\bibfield{author}{\bibinfo{person}{Florian Gallwitz} {and} \bibinfo{person}{Michael Kreil}.} \bibinfo{year}{2022}\natexlab{}.
\newblock \bibinfo{title}{Investigating the Validity of Botometer-based Social Bot Studies}.
\newblock
\newblock
\showeprint[arxiv]{2207.11474}~[cs.SI]


\bibitem[Garcia-Silva et~al\mbox{.}(2019)]%
        {garcia-silva2019an}
\bibfield{author}{\bibinfo{person}{Andres Garcia-Silva}, \bibinfo{person}{Cristian Berrio}, {and} \bibinfo{person}{José~Manuel Gómez-Pérez}.} \bibinfo{year}{2019}\natexlab{}.
\newblock \showarticletitle{An Empirical Study on Pre-trained Embeddings and Language Models for Bot Detection}. In \bibinfo{booktitle}{\emph{4th Workshop on Representation Learning for NLP (RepL4NLP)}}. \bibinfo{publisher}{Association for Computational Linguistics}, \bibinfo{pages}{148--155}.
\newblock
\urldef\tempurl%
\url{https://doi.org/10.18653/v1/W19-4317}
\showDOI{\tempurl}


\bibitem[Gebru et~al\mbox{.}(2021)]%
        {gebru2021datasheets}
\bibfield{author}{\bibinfo{person}{Timnit Gebru}, \bibinfo{person}{Jamie Morgenstern}, \bibinfo{person}{Briana Vecchione}, \bibinfo{person}{Jennifer~Wortman Vaughan}, \bibinfo{person}{Hanna Wallach}, \bibinfo{person}{Hal~Daum{\'e} Iii}, {and} \bibinfo{person}{Kate Crawford}.} \bibinfo{year}{2021}\natexlab{}.
\newblock \showarticletitle{Datasheets for datasets}.
\newblock \bibinfo{journal}{\emph{Commun. ACM}} \bibinfo{volume}{64}, \bibinfo{number}{12} (\bibinfo{year}{2021}), \bibinfo{pages}{86--92}.
\newblock


\bibitem[Gilani et~al\mbox{.}(2019)]%
        {gilani2019a}
\bibfield{author}{\bibinfo{person}{Zafar Gilani}, \bibinfo{person}{Reza Farahbakhsh}, \bibinfo{person}{Gareth Tyson}, {and} \bibinfo{person}{Jon Crowcroft}.} \bibinfo{year}{2019}\natexlab{}.
\newblock \showarticletitle{A Large-scale Behavioural Analysis of Bots and Humans on Twitter}.
\newblock \bibinfo{journal}{\emph{ACM Transactions on the Web}} \bibinfo{volume}{13}, \bibinfo{number}{1} (\bibinfo{year}{2019}).
\newblock
\urldef\tempurl%
\url{https://doi.org/10.1145/3298789}
\showDOI{\tempurl}


\bibitem[Gilani et~al\mbox{.}(2017a)]%
        {gilani2017of}
\bibfield{author}{\bibinfo{person}{Zafar Gilani}, \bibinfo{person}{Reza Farahbakhsh}, \bibinfo{person}{Gareth Tyson}, \bibinfo{person}{Liang Wang}, {and} \bibinfo{person}{Jon Crowcroft}.} \bibinfo{year}{2017}\natexlab{a}.
\newblock \showarticletitle{Of Bots and Humans (on Twitter)}. In \bibinfo{booktitle}{\emph{IEEE/ACM International Conference on Advances in Social Networks Analysis and Mining}}. \bibinfo{publisher}{Association for Computing Machinery}, \bibinfo{pages}{349--354}.
\newblock
\urldef\tempurl%
\url{https://doi.org/10.1145/3110025.3110090}
\showDOI{\tempurl}


\bibitem[Gilani et~al\mbox{.}(2017b)]%
        {gilani2017classification}
\bibfield{author}{\bibinfo{person}{Zafar Gilani}, \bibinfo{person}{Ekaterina Kochmar}, {and} \bibinfo{person}{Jon Crowcroft}.} \bibinfo{year}{2017}\natexlab{b}.
\newblock \showarticletitle{Classification of Twitter Accounts into Automated Agents and Human Users}. In \bibinfo{booktitle}{\emph{2017 IEEE/ACM International Conference on Advances in Social Networks Analysis and Mining (ASONAM)}}. \bibinfo{pages}{489--496}.
\newblock


\bibitem[Gilani et~al\mbox{.}(2020)]%
        {gilani2020classification}
\bibfield{author}{\bibinfo{person}{Zafar Gilani}, \bibinfo{person}{Ekaterina Kochmar}, {and} \bibinfo{person}{Jon Crowcroft}.} \bibinfo{year}{2020}\natexlab{}.
\newblock \showarticletitle{Classification of twitter accounts into automated agents and human users}. In \bibinfo{booktitle}{\emph{IEEE/ACM International Conference on Advances in Social Networks Analysis and Mining (ASONAM)}}. \bibinfo{pages}{489--496}.
\newblock
\urldef\tempurl%
\url{https://doi.org/10.1145/3110025.3110091}
\showDOI{\tempurl}


\bibitem[Giorgi et~al\mbox{.}(2021)]%
        {giorgi2021characterizing}
\bibfield{author}{\bibinfo{person}{Salvatore Giorgi}, \bibinfo{person}{Lyle Ungar}, {and} \bibinfo{person}{H.~Andrew Schwartz}.} \bibinfo{year}{2021}\natexlab{}.
\newblock \showarticletitle{Characterizing Social Spambots by their Human Traits}. In \bibinfo{booktitle}{\emph{The Joint Conference of the 59th Annual Meeting of the Association for Computational Linguistics and the 11th International Joint Conference on Natural Language Processing}}. \bibinfo{pages}{5148--5158}.
\newblock
\urldef\tempurl%
\url{https://doi.org/10.18653/v1/2021.findings-acl.457}
\showDOI{\tempurl}


\bibitem[González-Bailón et~al\mbox{.}(2022)]%
        {gonzalezbailon2022}
\bibfield{author}{\bibinfo{person}{Sandra González-Bailón}, \bibinfo{person}{Valeria d'Andrea}, \bibinfo{person}{Deen Freelon}, {and} \bibinfo{person}{Manlio De~Domenico}.} \bibinfo{year}{2022}\natexlab{}.
\newblock \showarticletitle{{The advantage of the right in social media news sharing}}.
\newblock \bibinfo{journal}{\emph{PNAS Nexus}} \bibinfo{volume}{1}, \bibinfo{number}{3} (\bibinfo{date}{07} \bibinfo{year}{2022}).
\newblock
\showISSN{2752-6542}
\urldef\tempurl%
\url{https://doi.org/10.1093/pnasnexus/pgac137}
\showDOI{\tempurl}
\showeprint{https://academic.oup.com/pnasnexus/article-pdf/1/3/pgac137/45484944/pgac137.pdf}
\newblock
\shownote{pgac137}.


\bibitem[Gorodnichenko et~al\mbox{.}(2018)]%
        {gorodnichenko2018social}
\bibfield{author}{\bibinfo{person}{Yuriy Gorodnichenko}, \bibinfo{person}{Tho Pham}, {and} \bibinfo{person}{Oleksandr Talavera}.} \bibinfo{year}{2018}\natexlab{}.
\newblock \bibinfo{booktitle}{\emph{{Social Media, Sentiment and Public Opinions: Evidence from \#Brexit and \#USElection}}}.
\newblock \bibinfo{type}{NBER Working Papers} 24631. \bibinfo{institution}{National Bureau of Economic Research, Inc}.
\newblock
\urldef\tempurl%
\url{https://ideas.repec.org/p/nbr/nberwo/24631.html}
\showURL{%
\tempurl}


\bibitem[Guo et~al\mbox{.}(2022)]%
        {guo2022social}
\bibfield{author}{\bibinfo{person}{Qinglang Guo}, \bibinfo{person}{Haiyong Xie}, \bibinfo{person}{Yangyang Li}, \bibinfo{person}{Wen Ma}, {and} \bibinfo{person}{Chao Zhang}.} \bibinfo{year}{2022}\natexlab{}.
\newblock \showarticletitle{Social Bots Detection via Fusing BERT and Graph Convolutional Networks}.
\newblock \bibinfo{journal}{\emph{Symmetry}} \bibinfo{volume}{14}, \bibinfo{number}{1} (\bibinfo{year}{2022}).
\newblock
\urldef\tempurl%
\url{https://doi.org/10.3390/sym14010030}
\showDOI{\tempurl}


\bibitem[Heidari and Jones(2020)]%
        {heidari2020using}
\bibfield{author}{\bibinfo{person}{Maryam Heidari} {and} \bibinfo{person}{James~H Jones}.} \bibinfo{year}{2020}\natexlab{}.
\newblock \showarticletitle{Using BERT to Extract Topic-Independent Sentiment Features for Social Media Bot Detection}. In \bibinfo{booktitle}{\emph{11th IEEE Annual Ubiquitous Computing, Electronics \& Mobile Communication Conference (UEMCON)}}. \bibinfo{pages}{542--547}.
\newblock
\urldef\tempurl%
\url{https://doi.org/10.1109/UEMCON51285.2020.9298158}
\showDOI{\tempurl}


\bibitem[Ilias and Roussaki(2021)]%
        {ilias2021detecting}
\bibfield{author}{\bibinfo{person}{Loukas Ilias} {and} \bibinfo{person}{Ioanna Roussaki}.} \bibinfo{year}{2021}\natexlab{}.
\newblock \showarticletitle{Detecting malicious activity in Twitter using deep learning techniques}.
\newblock \bibinfo{journal}{\emph{Applied Soft Computing}}  \bibinfo{volume}{107} (\bibinfo{year}{2021}).
\newblock


\bibitem[Jang et~al\mbox{.}(2018)]%
        {jang2018a}
\bibfield{author}{\bibinfo{person}{S.~Mo Jang}, \bibinfo{person}{Tieming Geng}, \bibinfo{person}{Jo-Yun~Queenie Li}, \bibinfo{person}{Ruofan Xia}, \bibinfo{person}{Chin-Tser Huang}, \bibinfo{person}{Hwalbin Kim}, {and} \bibinfo{person}{Jijun Tang}.} \bibinfo{year}{2018}\natexlab{}.
\newblock \showarticletitle{A computational approach for examining the roots and spreading patterns of fake news: Evolution tree analysis}.
\newblock \bibinfo{journal}{\emph{Computers in Human Behavior}}  \bibinfo{volume}{84} (\bibinfo{year}{2018}), \bibinfo{pages}{103--113}.
\newblock
\urldef\tempurl%
\url{https://doi.org/10.1016/j.chb.2018.02.032}
\showDOI{\tempurl}


\bibitem[Keller and Klinger(2018)]%
        {keller2018social}
\bibfield{author}{\bibinfo{person}{Tobias Keller} {and} \bibinfo{person}{Ulrike Klinger}.} \bibinfo{year}{2018}\natexlab{}.
\newblock \showarticletitle{Social Bots in Election Campaigns: Theoretical, Empirical, and Methodological Implications}.
\newblock \bibinfo{journal}{\emph{Political Communication}} \bibinfo{volume}{36}, \bibinfo{number}{1} (\bibinfo{year}{2018}), \bibinfo{pages}{171--189}.
\newblock
\urldef\tempurl%
\url{https://doi.org/10.1080/10584609.2018.1526238}
\showDOI{\tempurl}


\bibitem[Kouvela et~al\mbox{.}(2020)]%
        {kouvela2020bot}
\bibfield{author}{\bibinfo{person}{Maria Kouvela}, \bibinfo{person}{Ilias Dimitriadis}, {and} \bibinfo{person}{Athena Vakali}.} \bibinfo{year}{2020}\natexlab{}.
\newblock \showarticletitle{Bot-Detective: An Explainable Twitter Bot Detection Service with Crowdsourcing Functionalities}. In \bibinfo{booktitle}{\emph{Proceedings of the 12th International Conference on Management of Digital EcoSystems}} (Virtual Event, United Arab Emirates) \emph{(\bibinfo{series}{MEDES '20})}. \bibinfo{publisher}{Association for Computing Machinery}, \bibinfo{address}{New York, NY, USA}, \bibinfo{pages}{55–63}.
\newblock
\showISBNx{9781450381154}
\urldef\tempurl%
\url{https://doi.org/10.1145/3415958.3433075}
\showDOI{\tempurl}


\bibitem[Kudugunta and Ferrara(2018)]%
        {kudugunta2018deep}
\bibfield{author}{\bibinfo{person}{Sneha Kudugunta} {and} \bibinfo{person}{Emilio Ferrara}.} \bibinfo{year}{2018}\natexlab{}.
\newblock \showarticletitle{Deep neural networks for bot detection}.
\newblock \bibinfo{journal}{\emph{Information Sciences}}  \bibinfo{volume}{467} (\bibinfo{year}{2018}), \bibinfo{pages}{312--322}.
\newblock
\urldef\tempurl%
\url{https://doi.org/10.1016/j.ins.2018.08.019}
\showDOI{\tempurl}


\bibitem[Lee et~al\mbox{.}(2011)]%
        {lee2011a}
\bibfield{author}{\bibinfo{person}{Kyumin Lee}, \bibinfo{person}{Brian Eoff}, {and} \bibinfo{person}{James Caverlee}.} \bibinfo{year}{2011}\natexlab{}.
\newblock \showarticletitle{A Long-Term Study of Content Polluters on Twitter}. In \bibinfo{booktitle}{\emph{Fifth International AAAI Conference on Weblogs and Social Media}}. \bibinfo{publisher}{Association for the Advancement of Artificial Intelligence}, \bibinfo{pages}{185--192}.
\newblock


\bibitem[Lundberg and Lee(2017)]%
        {lundberg2017a}
\bibfield{author}{\bibinfo{person}{Scott~M Lundberg} {and} \bibinfo{person}{Su-In Lee}.} \bibinfo{year}{2017}\natexlab{}.
\newblock \showarticletitle{A Unified Approach to Interpreting Model Predictions}.
\newblock In \bibinfo{booktitle}{\emph{Advances in Neural Information Processing Systems 30}}, \bibfield{editor}{\bibinfo{person}{I.~Guyon}, \bibinfo{person}{U.~V. Luxburg}, \bibinfo{person}{S.~Bengio}, \bibinfo{person}{H.~Wallach}, \bibinfo{person}{R.~Fergus}, \bibinfo{person}{S.~Vishwanathan}, {and} \bibinfo{person}{R.~Garnett}} (Eds.). \bibinfo{publisher}{Curran Associates, Inc.}, \bibinfo{pages}{4765--4774}.
\newblock
\urldef\tempurl%
\url{http://papers.nips.cc/paper/7062-a-unified-approach-to-interpreting-model-predictions.pdf}
\showURL{%
\tempurl}


\bibitem[Luo et~al\mbox{.}(2019)]%
        {luo2019deepbot}
\bibfield{author}{\bibinfo{person}{Linhao Luo}, \bibinfo{person}{Xiaofeng Zhang}, \bibinfo{person}{Xiaofei Yang}, {and} \bibinfo{person}{Weihuang Yang}.} \bibinfo{year}{2019}\natexlab{}.
\newblock \showarticletitle{Deepbot: A Deep Neural Network based approach for Detecting Twitter Bots}.
\newblock \bibinfo{journal}{\emph{{IOP} Conference Series: Materials Science and Engineering}} \bibinfo{volume}{719}, \bibinfo{number}{1} (\bibinfo{year}{2019}).
\newblock
\urldef\tempurl%
\url{https://doi.org/10.1088/1757-899x/719/1/012063}
\showDOI{\tempurl}


\bibitem[Martini et~al\mbox{.}(2021)]%
        {martini2021bot}
\bibfield{author}{\bibinfo{person}{Franziska Martini}, \bibinfo{person}{Paul Samula}, \bibinfo{person}{Tobias~R Keller}, {and} \bibinfo{person}{Ulrike Klinger}.} \bibinfo{year}{2021}\natexlab{}.
\newblock \showarticletitle{Bot, or not? Comparing three methods for detecting social bots in five political discourses}.
\newblock \bibinfo{journal}{\emph{Big Data \& Society}} \bibinfo{volume}{8}, \bibinfo{number}{2} (\bibinfo{year}{2021}), \bibinfo{pages}{20539517211033566}.
\newblock
\urldef\tempurl%
\url{https://doi.org/10.1177/20539517211033566}
\showDOI{\tempurl}
\showeprint{https://doi.org/10.1177/20539517211033566}


\bibitem[Martín-Gutiérrez et~al\mbox{.}(2021)]%
        {ilias2021deep}
\bibfield{author}{\bibinfo{person}{David Martín-Gutiérrez}, \bibinfo{person}{Gustavo Hernández-Peñaloza}, \bibinfo{person}{Alberto~Belmonte Hernández}, \bibinfo{person}{Alicia Lozano-Diez}, {and} \bibinfo{person}{Federico Álvarez}.} \bibinfo{year}{2021}\natexlab{}.
\newblock \showarticletitle{A Deep Learning Approach for Robust Detection of Bots in Twitter Using Transformers}.
\newblock \bibinfo{journal}{\emph{IEEE Access}}  \bibinfo{volume}{9} (\bibinfo{year}{2021}), \bibinfo{pages}{54591--54601}.
\newblock
\urldef\tempurl%
\url{https://doi.org/10.1109/ACCESS.2021.3068659}
\showDOI{\tempurl}


\bibitem[Mazza et~al\mbox{.}(2019)]%
        {mazza2019rtbust}
\bibfield{author}{\bibinfo{person}{Michele Mazza}, \bibinfo{person}{Stefano Cresci}, \bibinfo{person}{Marco Avvenuti}, \bibinfo{person}{Walter Quattrociocchi}, {and} \bibinfo{person}{Maurizio Tesconi}.} \bibinfo{year}{2019}\natexlab{}.
\newblock \showarticletitle{RTbust: Exploiting Temporal Patterns for Botnet Detection on Twitter}. In \bibinfo{booktitle}{\emph{10th ACM Conference on Web Science}}. \bibinfo{publisher}{Association for Computing Machinery}, \bibinfo{pages}{183--192}.
\newblock
\urldef\tempurl%
\url{https://doi.org/10.1145/3292522.3326015}
\showDOI{\tempurl}


\bibitem[Mou and Lee(2020)]%
        {mou2020malicious}
\bibfield{author}{\bibinfo{person}{Guanyi Mou} {and} \bibinfo{person}{Kyumin Lee}.} \bibinfo{year}{2020}\natexlab{}.
\newblock \showarticletitle{Malicious bot detection in online social networks: arming handcrafted features with deep learning}. In \bibinfo{booktitle}{\emph{International Conference on Social Informatics}}. Springer, \bibinfo{pages}{220--236}.
\newblock


\bibitem[Musk(2022)]%
        {musk2022mdau}
\bibfield{author}{\bibinfo{person}{Elon Musk}.} \bibinfo{year}{2022}\natexlab{}.
\newblock \bibinfo{title}{Bot Percentage Thread}.
\newblock
\newblock
\urldef\tempurl%
\url{https://twitter.com/elonmusk/status/1555950698252181507}
\showURL{%
\tempurl}


\bibitem[Nizzoli et~al\mbox{.}(2020)]%
        {nizzoli2020access}
\bibfield{author}{\bibinfo{person}{Leonardo Nizzoli}, \bibinfo{person}{Serena Tardelli}, \bibinfo{person}{Marco Avvenuti}, \bibinfo{person}{Stefano Cresci}, \bibinfo{person}{Maurizio Tesconi}, {and} \bibinfo{person}{Emilio Ferrara}.} \bibinfo{year}{2020}\natexlab{}.
\newblock \showarticletitle{Charting the Landscape of Online Cryptocurrency Manipulation}.
\newblock \bibinfo{journal}{\emph{IEEE Access}}  \bibinfo{volume}{8} (\bibinfo{year}{2020}), \bibinfo{pages}{113230--113245}.
\newblock
\urldef\tempurl%
\url{https://doi.org/10.1109/ACCESS.2020.3003370}
\showDOI{\tempurl}


\bibitem[Pennycook et~al\mbox{.}(2021)]%
        {pennycook2021shifting}
\bibfield{author}{\bibinfo{person}{Gordon Pennycook}, \bibinfo{person}{Ziv Epstein}, \bibinfo{person}{Mohsen Mosleh}, \bibinfo{person}{Antonio~A. Arechar}, \bibinfo{person}{Dean Eckles}, {and} \bibinfo{person}{David~G. Rand}.} \bibinfo{year}{2021}\natexlab{}.
\newblock \showarticletitle{Shifting attention to accuracy can reduce misinformation online}.
\newblock \bibinfo{journal}{\emph{Nature}} \bibinfo{volume}{592}, \bibinfo{number}{6380} (\bibinfo{year}{2021}), \bibinfo{pages}{590--595}.
\newblock
\urldef\tempurl%
\url{https://doi.org/10.1038/s41586-021-03344-2}
\showDOI{\tempurl}


\bibitem[Pierri et~al\mbox{.}(2020)]%
        {pierri2020investigating}
\bibfield{author}{\bibinfo{person}{Francesco Pierri}, \bibinfo{person}{Alessandro Artoni}, {and} \bibinfo{person}{Stefano Ceri}.} \bibinfo{year}{2020}\natexlab{}.
\newblock \showarticletitle{Investigating Italian disinformation spreading on Twitter in the context of 2019 European elections}.
\newblock \bibinfo{journal}{\emph{PloS one}} \bibinfo{volume}{15}, \bibinfo{number}{1} (\bibinfo{year}{2020}), \bibinfo{pages}{e0227821}.
\newblock
\urldef\tempurl%
\url{https://doi.org/10.1371/journal.pone.0227821}
\showDOI{\tempurl}


\bibitem[Rangel and Rosso(2015)]%
        {rangel2015overview}
\bibfield{author}{\bibinfo{person}{Francisco Rangel} {and} \bibinfo{person}{Paolo Rosso}.} \bibinfo{year}{2015}\natexlab{}.
\newblock \showarticletitle{Overview of the 7th Author Profiling Task at PAN 2019: Bots and Gender Profiling in Twitter}. In \bibinfo{booktitle}{\emph{CLEF Evaluation Labs and Workshop Working Notes Papers}}.
\newblock


\bibitem[Rauchfleisch and Kaiser(2020)]%
        {rauchfleisch2020false}
\bibfield{author}{\bibinfo{person}{Adrian Rauchfleisch} {and} \bibinfo{person}{Jonas Kaiser}.} \bibinfo{year}{2020}\natexlab{}.
\newblock \showarticletitle{The False positive problem of automatic bot detection in social science research}.
\newblock \bibinfo{journal}{\emph{PLOS ONE}} \bibinfo{volume}{15}, \bibinfo{number}{10} (\bibinfo{date}{10} \bibinfo{year}{2020}), \bibinfo{pages}{1--20}.
\newblock
\urldef\tempurl%
\url{https://doi.org/10.1371/journal.pone.0241045}
\showDOI{\tempurl}


\bibitem[Ribeiro et~al\mbox{.}(2016)]%
        {ribeiro2016why}
\bibfield{author}{\bibinfo{person}{Marco~Tulio Ribeiro}, \bibinfo{person}{Sameer Singh}, {and} \bibinfo{person}{Carlos Guestrin}.} \bibinfo{year}{2016}\natexlab{}.
\newblock \showarticletitle{"Why Should I Trust You?": Explaining the Predictions of Any Classifier}. In \bibinfo{booktitle}{\emph{Proceedings of the 22nd ACM SIGKDD International Conference on Knowledge Discovery and Data Mining}} (San Francisco, California, USA) \emph{(\bibinfo{series}{KDD '16})}. \bibinfo{publisher}{Association for Computing Machinery}, \bibinfo{address}{New York, NY, USA}, \bibinfo{pages}{1135–1144}.
\newblock
\showISBNx{9781450342322}
\urldef\tempurl%
\url{https://doi.org/10.1145/2939672.2939778}
\showDOI{\tempurl}


\bibitem[Rovito et~al\mbox{.}(2022)]%
        {rovito2022an}
\bibfield{author}{\bibinfo{person}{Luigi Rovito}, \bibinfo{person}{Lorenzo Bonin}, \bibinfo{person}{Luca Manzoni}, {and} \bibinfo{person}{Andrea De~Lorenzo}.} \bibinfo{year}{2022}\natexlab{}.
\newblock \showarticletitle{An Evolutionary Computation Approach for Twitter Bot Detection}.
\newblock \bibinfo{journal}{\emph{Applied Sciences}} \bibinfo{volume}{12}, \bibinfo{number}{12} (\bibinfo{year}{2022}).
\newblock
\urldef\tempurl%
\url{https://doi.org/10.3390/app12125915}
\showDOI{\tempurl}


\bibitem[Rudin(2019)]%
        {rudin2019stop}
\bibfield{author}{\bibinfo{person}{Cynthia Rudin}.} \bibinfo{year}{2019}\natexlab{}.
\newblock \showarticletitle{Stop explaining black box machine learning models for high stakes decisions and use interpretable models instead}.
\newblock \bibinfo{journal}{\emph{Nature Machine Intelligence}} \bibinfo{volume}{1}, \bibinfo{number}{5} (\bibinfo{date}{01 May} \bibinfo{year}{2019}), \bibinfo{pages}{206--215}.
\newblock
\showISSN{2522-5839}
\urldef\tempurl%
\url{https://doi.org/10.1038/s42256-019-0048-x}
\showDOI{\tempurl}


\bibitem[Sayyadiharikandeh et~al\mbox{.}(2020)]%
        {sayyadiharikandeh2020detection}
\bibfield{author}{\bibinfo{person}{Mohsen Sayyadiharikandeh}, \bibinfo{person}{Onur Varol}, \bibinfo{person}{Kai-Cheng Yang}, \bibinfo{person}{Alessandro Flammini}, {and} \bibinfo{person}{Filippo Menczer}.} \bibinfo{year}{2020}\natexlab{}.
\newblock \showarticletitle{Detection of Novel Social Bots by Ensembles of Specialized Classifiers}. In \bibinfo{booktitle}{\emph{29th ACM International Conference on Information \& Knowledge Management}}. \bibinfo{publisher}{Association for Computing Machinery}, \bibinfo{pages}{2725--2732}.
\newblock
\urldef\tempurl%
\url{https://doi.org/10.1145/3340531.3412698}
\showDOI{\tempurl}


\bibitem[Shao et~al\mbox{.}(2018a)]%
        {shao2018the}
\bibfield{author}{\bibinfo{person}{Chengcheng Shao}, \bibinfo{person}{Giovanni~Luca Ciampaglia}, \bibinfo{person}{Onur Varol}, \bibinfo{person}{Kai-Cheng Yang}, \bibinfo{person}{Alessandro Flammini}, {and} \bibinfo{person}{Filippo Menczer}.} \bibinfo{year}{2018}\natexlab{a}.
\newblock \showarticletitle{The spread of low-credibility content by social bots}.
\newblock \bibinfo{journal}{\emph{Nature Communications}} \bibinfo{volume}{9}, \bibinfo{number}{4787} (\bibinfo{year}{2018}), \bibinfo{pages}{1--9}.
\newblock
\urldef\tempurl%
\url{https://doi.org/10.1038/s41467-018-06930-7}
\showDOI{\tempurl}


\bibitem[Shao et~al\mbox{.}(2018b)]%
        {shao2018anatomy}
\bibfield{author}{\bibinfo{person}{Chengcheng Shao}, \bibinfo{person}{Pik-Mai Hui}, \bibinfo{person}{Lei Wang}, \bibinfo{person}{Xinwen Jiang}, \bibinfo{person}{Alessandro Flammini}, \bibinfo{person}{Filippo Menczer}, {and} \bibinfo{person}{Giovanni~Luca Ciampaglia}.} \bibinfo{year}{2018}\natexlab{b}.
\newblock \showarticletitle{Anatomy of an online misinformation network}.
\newblock \bibinfo{journal}{\emph{PloS one}} \bibinfo{volume}{13}, \bibinfo{number}{4} (\bibinfo{year}{2018}), \bibinfo{pages}{e0196087}.
\newblock
\urldef\tempurl%
\url{https://doi.org/10.1371/journal.pone.0196087}
\showDOI{\tempurl}


\bibitem[Shu et~al\mbox{.}(2020)]%
        {shu2020fakenewsnet:}
\bibfield{author}{\bibinfo{person}{Kai Shu}, \bibinfo{person}{Deepak Mahudeswaran}, \bibinfo{person}{Suhang Wang}, \bibinfo{person}{Dongwon Lee}, {and} \bibinfo{person}{Huan Liu}.} \bibinfo{year}{2020}\natexlab{}.
\newblock \showarticletitle{FakeNewsNet: A Data Repository with News Content, Social Context, and Spatiotemporal Information for Studying Fake News on Social Media}.
\newblock \bibinfo{journal}{\emph{Big Data}} \bibinfo{volume}{8}, \bibinfo{number}{3} (\bibinfo{year}{2020}), \bibinfo{pages}{171--188}.
\newblock
\urldef\tempurl%
\url{https://doi.org/10.1089/big.2020.0062}
\showDOI{\tempurl}


\bibitem[Stella et~al\mbox{.}(2018)]%
        {stella2018bots}
\bibfield{author}{\bibinfo{person}{Massimo Stella}, \bibinfo{person}{Emilio Ferrara}, {and} \bibinfo{person}{Manlio~De Domenico}.} \bibinfo{year}{2018}\natexlab{}.
\newblock \showarticletitle{Bots increase exposure to negative and inflammatory content in online social systems}.
\newblock \bibinfo{journal}{\emph{Proceedings of the National Academy of Sciences}} \bibinfo{volume}{115}, \bibinfo{number}{49} (\bibinfo{year}{2018}), \bibinfo{pages}{12435--12440}.
\newblock
\urldef\tempurl%
\url{https://doi.org/10.1073/pnas.1803470115}
\showDOI{\tempurl}


\bibitem[Torusdağ et~al\mbox{.}(2020)]%
        {torusdag2020secure}
\bibfield{author}{\bibinfo{person}{M.~Buğra Torusdağ}, \bibinfo{person}{Mucahid Kutlu}, {and} \bibinfo{person}{Ali~Aydın Selçuk}.} \bibinfo{year}{2020}\natexlab{}.
\newblock \showarticletitle{Are We Secure from Bots? Investigating Vulnerabilities of Botometer}. In \bibinfo{booktitle}{\emph{2020 5th International Conference on Computer Science and Engineering (UBMK)}}. \bibinfo{pages}{343--348}.
\newblock
\urldef\tempurl%
\url{https://doi.org/10.1109/UBMK50275.2020.9219433}
\showDOI{\tempurl}


\bibitem[Twitter(2021)]%
        {twitter2021mdau}
\bibfield{author}{\bibinfo{person}{Twitter}.} \bibinfo{year}{2021}\natexlab{}.
\newblock \bibinfo{title}{FORM 10-K}.
\newblock
\newblock
\urldef\tempurl%
\url{https://www.sec.gov/Archives/edgar/data/1418091/000141809121000031/twtr-20201231.htm}
\showURL{%
\tempurl}


\bibitem[Vosoughi et~al\mbox{.}(2018)]%
        {vosoughi2018spread}
\bibfield{author}{\bibinfo{person}{Soroush Vosoughi}, \bibinfo{person}{Deb Roy}, {and} \bibinfo{person}{Sinan Aral}.} \bibinfo{year}{2018}\natexlab{}.
\newblock \showarticletitle{The spread of true and false news online}.
\newblock \bibinfo{journal}{\emph{Science}} \bibinfo{volume}{359}, \bibinfo{number}{6380} (\bibinfo{year}{2018}), \bibinfo{pages}{1146--1151}.
\newblock
\urldef\tempurl%
\url{https://doi.org/10.1126/science.aap9559}
\showDOI{\tempurl}
\showeprint{https://www.science.org/doi/pdf/10.1126/science.aap9559}


\bibitem[Wang et~al\mbox{.}(2021)]%
        {wang2021detecting}
\bibfield{author}{\bibinfo{person}{Xiujuan Wang}, \bibinfo{person}{Qianqian Zheng}, \bibinfo{person}{Kangfeng Zheng}, \bibinfo{person}{Yi Sui}, \bibinfo{person}{Siwei Cao}, {and} \bibinfo{person}{Yutong Shi}.} \bibinfo{year}{2021}\natexlab{}.
\newblock \showarticletitle{Detecting social media bots with variational autoencoder and k-nearest neighbor}.
\newblock \bibinfo{journal}{\emph{Applied Sciences}} \bibinfo{volume}{11}, \bibinfo{number}{12} (\bibinfo{year}{2021}), \bibinfo{pages}{5482}.
\newblock


\bibitem[Wischnewski et~al\mbox{.}(2021)]%
        {wischnewski2021disagree}
\bibfield{author}{\bibinfo{person}{Magdalena Wischnewski}, \bibinfo{person}{Rebecca Bernemann}, \bibinfo{person}{Thao Ngo}, {and} \bibinfo{person}{Nicole Kr\"{a}mer}.} \bibinfo{year}{2021}\natexlab{}.
\newblock \showarticletitle{Disagree? You Must Be a Bot! How Beliefs Shape Twitter Profile Perceptions}. In \bibinfo{booktitle}{\emph{Proceedings of the 2021 CHI Conference on Human Factors in Computing Systems}} (Yokohama, Japan) \emph{(\bibinfo{series}{CHI '21})}. \bibinfo{publisher}{Association for Computing Machinery}, \bibinfo{address}{New York, NY, USA}, Article \bibinfo{articleno}{160}, \bibinfo{numpages}{11}~pages.
\newblock
\showISBNx{9781450380966}
\urldef\tempurl%
\url{https://doi.org/10.1145/3411764.3445109}
\showDOI{\tempurl}


\bibitem[Wu et~al\mbox{.}(2017)]%
        {wu2017adaptive}
\bibfield{author}{\bibinfo{person}{Liang Wu}, \bibinfo{person}{Xia Hu}, \bibinfo{person}{Fred Morstatter}, {and} \bibinfo{person}{Huan Liu}.} \bibinfo{year}{2017}\natexlab{}.
\newblock \showarticletitle{Adaptive Spammer Detection with Sparse Group Modeling}. In \bibinfo{booktitle}{\emph{Eleventh International AAAI Conference on Web and Social Media (ICWSM)}}. \bibinfo{publisher}{Association for the Advancement of Artificial Intelligence}, \bibinfo{pages}{319--326}.
\newblock


\bibitem[Yan et~al\mbox{.}(2021)]%
        {yan2021asymmetrical}
\bibfield{author}{\bibinfo{person}{Harry~Yaojun Yan}, \bibinfo{person}{Kai-Cheng Yang}, \bibinfo{person}{Filippo Menczer}, {and} \bibinfo{person}{James Shanahan}.} \bibinfo{year}{2021}\natexlab{}.
\newblock \showarticletitle{Asymmetrical perceptions of partisan political bots}.
\newblock \bibinfo{journal}{\emph{New Media \& Society}} \bibinfo{volume}{23}, \bibinfo{number}{10} (\bibinfo{year}{2021}), \bibinfo{pages}{3016--3037}.
\newblock
\urldef\tempurl%
\url{https://doi.org/10.1177/1461444820942744}
\showDOI{\tempurl}
\showeprint{https://doi.org/10.1177/1461444820942744}


\bibitem[Yang et~al\mbox{.}(2013)]%
        {yang2013empirical}
\bibfield{author}{\bibinfo{person}{Chao Yang}, \bibinfo{person}{Robert Harkreader}, {and} \bibinfo{person}{Guofei Gu}.} \bibinfo{year}{2013}\natexlab{}.
\newblock \showarticletitle{Empirical evaluation and new design for fighting evolving twitter spammers.}
\newblock \bibinfo{journal}{\emph{IEEE Transactions on Information Forensics and Security}} \bibinfo{volume}{8}, \bibinfo{number}{8} (\bibinfo{year}{2013}), \bibinfo{pages}{1280--1293}.
\newblock
\urldef\tempurl%
\url{https://doi.org/10.1109/TIFS.2013.2267732}
\showDOI{\tempurl}


\bibitem[Yang et~al\mbox{.}(2019)]%
        {yang2019arming}
\bibfield{author}{\bibinfo{person}{Kai‐Cheng Yang}, \bibinfo{person}{Onur Varol}, \bibinfo{person}{Clayton~A. Davis}, \bibinfo{person}{Emilio Ferrara}, \bibinfo{person}{Alessandro Flammini}, \bibinfo{person}{}, {and} \bibinfo{person}{Filippo Menczer}.} \bibinfo{year}{2019}\natexlab{}.
\newblock \showarticletitle{Arming the public with artificial intelligence to counter social bots.}
\newblock \bibinfo{journal}{\emph{Human Behavior and Emerging Technologies}}  \bibinfo{volume}{1} (\bibinfo{year}{2019}), \bibinfo{pages}{48--68}.
\newblock
\urldef\tempurl%
\url{https://doi.org/10.1002/hbe2.115}
\showDOI{\tempurl}


\bibitem[Yang et~al\mbox{.}(2022)]%
        {yang2022botometer101}
\bibfield{author}{\bibinfo{person}{Kai-Cheng Yang}, \bibinfo{person}{Emilio Ferrara}, {and} \bibinfo{person}{Filippo Menczer}.} \bibinfo{year}{2022}\natexlab{}.
\newblock \showarticletitle{Botometer 101: Social bot practicum for computational social scientists}.
\newblock \bibinfo{journal}{\emph{arXiv preprint arXiv:2201.01608}} (\bibinfo{year}{2022}).
\newblock


\bibitem[Yang et~al\mbox{.}(2020)]%
        {yang2020scalable}
\bibfield{author}{\bibinfo{person}{Kai-Cheng Yang}, \bibinfo{person}{Onur Varol}, \bibinfo{person}{Pik-Mai Hui}, {and} \bibinfo{person}{Filippo Menczer}.} \bibinfo{year}{2020}\natexlab{}.
\newblock \showarticletitle{Scalable and Generalizable Social Bot Detection through Data Selection}. In \bibinfo{booktitle}{\emph{AAAI Conference on Artificial Intelligence}}. \bibinfo{publisher}{Association for the Advancement of Artificial Intelligence}, \bibinfo{pages}{1096--1103}.
\newblock
\urldef\tempurl%
\url{https://doi.org/10.1609/aaai.v34i01.5460}
\showDOI{\tempurl}


\end{thebibliography}

\appendix

\end{document}